\pdfoutput=1

\documentclass[11pt]{article}

\usepackage[final]{acl}

\usepackage{times}
\usepackage{latexsym}

\usepackage{tabularray}
\usepackage{color}
\usepackage{multirow}
\usepackage{graphicx}
\usepackage{amsmath}
\usepackage{booktabs}
\usepackage{makecell}
\usepackage{pifont}
\usepackage{fontawesome}
\usepackage{graphicx}

\definecolor{darkgreen}{RGB}{48,128,20}
\definecolor{darkpink}{RGB}{255,153,204}
\definecolor{skyblue}{RGB}{51,102,255}

\usepackage[T1]{fontenc}

\usepackage[utf8]{inputenc}

\usepackage{xspace}
\newcommand{\ourdata}{ICLEval\xspace}

\usepackage{microtype}

\title{\ourdata: Evaluating In-Context Learning Ability of Large \\Language Models}

\author{
\bf Wentong Chen$^1$, Yankai Lin$^1$\thanks{\ \ Corresponding Author},  ZhenHao Zhou$^2$, HongYun Huang$^2$, \\
\bf Yantao Jia$^2$, Zhao Cao$^2$, Ji-Rong Wen$^{1,3,4}$ \\
  $^1$Gaoling School of Artificial Intelligence, Renmin University of China\\
  $^2$Huawei Poisson Lab \\
  $^3$Beijing Key Laboratory of Big Data Management and Analysis Methods \\
  $^4$School of Information, Renmin University of China
}

\begin{document}
\maketitle
\begin{abstract}
In-Context Learning (ICL) is a critical capability of Large Language Models (LLMs) as it empowers them to comprehend and reason across interconnected inputs. Evaluating the ICL ability of LLMs can enhance their utilization and deepen our understanding of how this ability is acquired at the training stage. However, existing evaluation frameworks primarily focus on language abilities and knowledge, often overlooking the assessment of ICL ability. In this work, we introduce the \ourdata benchmark to evaluate the ICL abilities of LLMs, which encompasses two key sub-abilities: exact copying and rule learning. Through the \ourdata benchmark, we demonstrate that ICL ability is universally present in different LLMs, and model size is not the sole determinant of ICL efficacy. Surprisingly, we observe that ICL abilities, particularly copying, develop early in the pretraining process and stabilize afterward. Our source codes and benchmark are released at \url{https://github.com/RUCBM/ICLEval}.

\end{abstract}

\section{Introduction}
Large Language Models (LLMs) possess stronger language abilities and richer knowledge compared to traditional models, and one of their most important differentiating factors is their In-Context Learning (ICL) ability~\citep{brown2020language}. ICL ability serves as a hallmark of LLMs and plays a crucial role in their strong generalization performance. It enables LLMs to quickly adapt to new tasks without altering their internal parameters, utilizing techniques like zero-shot and few-shot learning. Moreover, ICL is a pivotal component in tool learning \citep{qin2023tool} and various reasoning strategies, such as chain-of-thought ~\citep{wei2022chain} and tree-of-thought \citep{yao2023tree}.

The advantages of ICL have attracted many researchers to explore how and why it works \cite{lu2021fantastically, nie2022improving, ye2022complementary, min2022rethinking, liu2023lost}. Some researchers \cite{olsson2022context} find that induction heads can implement pattern copying behavior and appear to be the primary source of ICL. Pattern copying behavior means LLMs can exactly copy the previous content based on prefix matching or generate similar content based on learned rules. A natural idea is to do a quantitative evaluation including both aspects, to help better explore the sources and influencing factors of LLMs' ICL ability. Currently, evaluations of LLMs primarily focus on language abilities \citep{reddy2019coqa, lai2017race, rajpurkar2018know}, knowledge applications \citep{clark2018think, bisk2020piqa, zellers2019hellaswag, mihaylov2018can}, and complex abilities \citep{yuan2023well, chen2021evaluating, hendrycks2020measuring, huang2023c}. Utilizing these data as ICL tasks may introduce interference and hinder the assessment of LLMs' real ICL abilities.

In this work, we introduce the \ourdata benchmark to systematically evaluate the ICL ability of LLMs. Given that pattern copying behavior can be regarded as the main performance of ICL, we construct our tasks from two aspects: exact copying and rule learning. Exact copying means LLMs match the same prefix and copy the subsequent content, while rule learning means LLMs learn the rule from examples and complete similar content. We incorporate two scenarios (unstructured and structured contexts) to evaluate exact copying ability and consider various scenarios of rules (format, order, statistics, etc.) to assess rule learning ability.

We conduct experiments to explore the factors that affect ICL ability. 
Firstly, we consider if the \textbf{model size} is the key factor that affects the ICL ability. We test a series of LLMs ranging from 1.1B to 65B parameters. We observe that larger models often exhibit stronger ICL ability, but some smaller models can also be compared with the larger ones.
Secondly, we explore how the ICL ability changes in the \textbf{pretraining stage} by testing the checkpoints with different training tokens. Surprisingly, we find that most ICL abilities reach their peak in the early stages of pretraining, with minimal growth during subsequent training. Finally, we analyze some bad cases from four aspects.

\section{Benchmark Construction}

In-context learning is an inherent capability of LLMs, enabling them to perform various tasks without modifying their parameters. However, evaluating ICL directly presents challenges as it can be easily influenced by language proficiency and knowledge. For instance, comparing the ICL ability of two different models on common tasks like the GLUE benchmark \cite{wang2018glue} is challenging, even when using the same n-shot settings. If one model obtains a lower score, it could be attributed to a weakness in ICL ability or language understanding, making it difficult to determine the exact cause.
In this study, we aim to create evaluation scenarios for ICL that mitigate potential confounding factors.

\begin{table}[t!]
    \centering
    \small
    \begin{tabular}{llr}
        \toprule
        \textbf{Type} & \textbf{Task Name} & \textbf{Num.} \\
        \midrule
        \textbf{Exact Copying} & & \\
        Unstructured Text & String Completion & 100\\
        Structured Text & Dictionary Search &  190\\
        \midrule
        \textbf{Rule Learning} & & \\
        \multirow{3}{*}{Format Rules} &Format Check &  120 \\
        & Format Cloning & 100\\
        & Format Conversion & 120\\
        
        \multirow{2}{*}{Order Rules} &Order Check & 100\\
        & Order Adjustment & 240 \\
        
        \multirow{4}{*}{Statistics Rules} & Duplication Check  & 300\\
        & De-Duplication  & 300 \\
        & Count \& Navigation  & 120\\
        & Relation Analysis & 100\\

        List Mapping & Numbers' List Mapping &  250\\
        \midrule
        \textbf{Total} & & 2,040\\
        \bottomrule
    \end{tabular}
    \caption{The number of samples in our ICL tasks.}
    \label{tab:tasks-sample-number}
\end{table}   

\begin{figure*}[!t]
    \centering
    \includegraphics[width=1.0\linewidth]{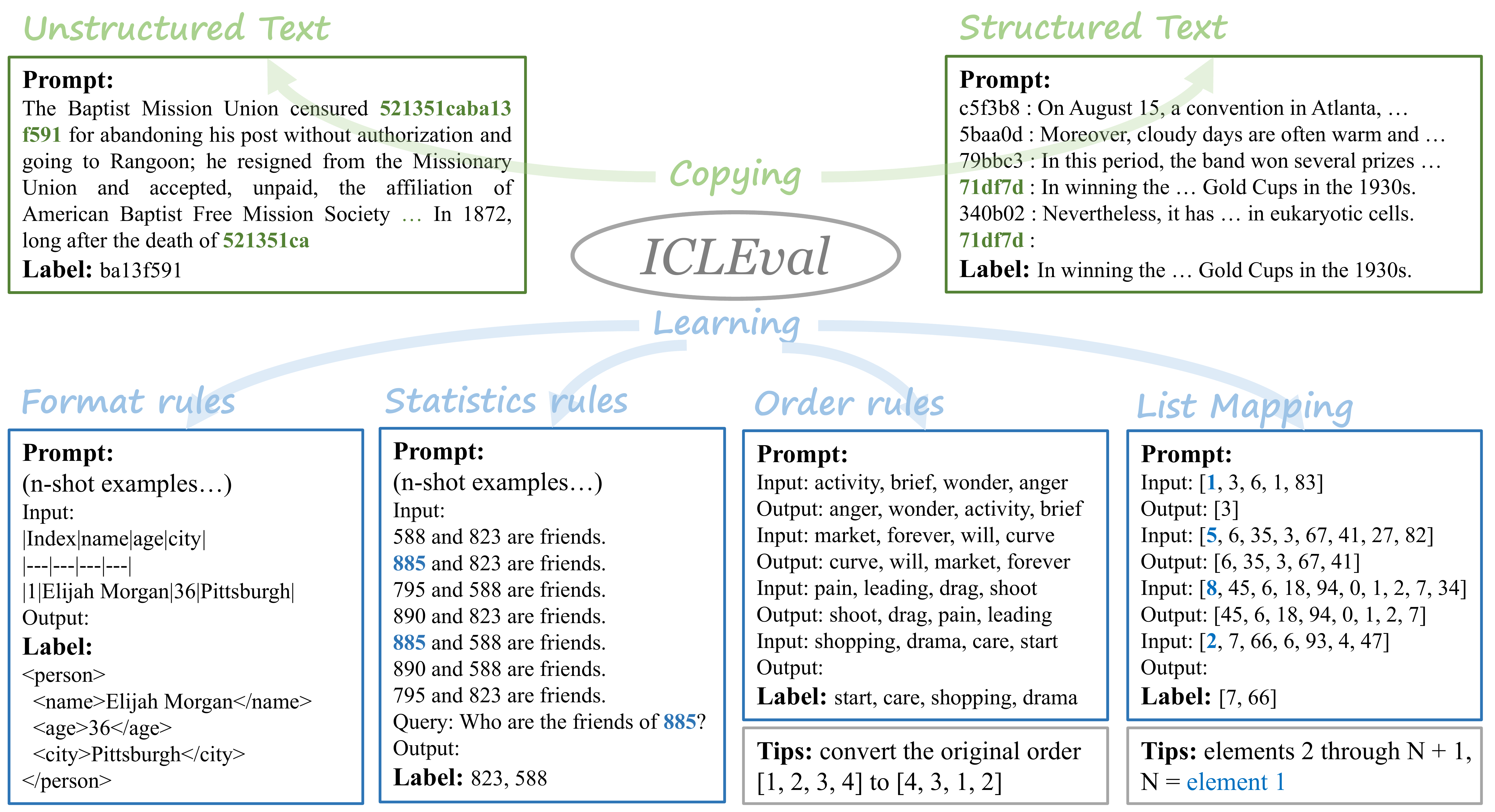}
    \caption{The examples of six representative tasks in \ourdata. }
    \label{fig:main}
\end{figure*}

ICL primarily hinges on two foundational skills: (1) the exact copying ability, which can be represented as "$AB \dots A \to B$" \, models need to perform two steps: first discriminate the "$A$" part, as referred to prefix matching and then copy the "$B$" part. (2) the rule learning ability, which enables the model to identify patterns and rules within the aggregated information derived from the n-shot examples, which are relevant to the task at hand. Therefore, we design several tasks for evaluating the copy and learning abilities in \ourdata. The categories and statistics for each task are presented in Table~\ref{tab:tasks-sample-number}, and more details about samples can see Appendix~\ref{sec:dataset samples}. 

\subsection{Exact Copying Ability}

Copying is a fundamental ability of the ICL mechanism, allowing LLMs to gather supplementary information from contextual cues. In our evaluation, we design tasks specifically to assess the exact copying ability of LLMs, wherein they are required to copy fragments that are present within the given context. To evaluate the models' adaptability and proficiency across various contextual scenarios, \ourdata categorizes copying tasks into two distinct types: unstructured (natural language) text and structured text.

\paragraph{Copying in unstructured text.} 
As shown in the left-top example of Figure~\ref{fig:main}, we mask the second half of a string (str. "521351caba13f591") which appears one or multiple times in the previous paragraphs. Then we require the models to predict it given the first half (str. "521351ca"). Notably, we use the hash strings to replace the origin entities in order to prevent LLMs from completing the entity based on their internal knowledge rather than the context.

\paragraph{Copying in structured text.}  
We use the "dictionary" format as a representation of structured data to better control the length and similarity of "$A$" part and "$B$" part.
As shown in the right-top example of Figure~\ref{fig:main}, this task challenges the models to deduce a value from a specified key (str. "71df7d") from a set of key-value pairs. We test their ability to seek and extract information from structured repositories efficiently.

\subsection{Rule Learning Ability}

The learning ability is another fundamental ability of the ICL mechanism, allowing LLMs to extract mapping rules from in-context examples. This ability enables them to tackle different and unseen tasks based on examples or natural language descriptions, without requiring updates to model parameters.  To focus solely on the learning ability and avoid the influence of language ability and knowledge across different models, \ourdata decomposes the evaluation of learning ability into four foundational aspects: format rules, order rules, statistics rules, and list mapping.

\textbf{Learning format rules} assesses whether LLMs can learn formatting mappings from in-context examples, a key skill for generating appropriately formatted outputs across various tasks. Specifically, we design a format classification task and two format generation tasks named format check, format cloning, and format conversation respectively. Figure~\ref{fig:main} shows an example of the format conversation task that converts "Markdown-Table" format data to "XML" format.

\textbf{Learning order rules} examines LLMs' ability to grasp the order of a group of elements as well as the mapping rule of two groups of elements from in-context examples. This skill is crucial for tasks requiring re-organization of input elements, such as in translation and syntactic analysis scenarios. To this end, we formulate a classification task and a generation task aimed at evaluating LLMs' capabilities in determining whether inputs satisfy a specific order (named order check) and in executing order transformations (named order adjustment). Figure~\ref{fig:main} shows an example of the order adjustment task that converts the word-level order from $[1,2,3,4]$ to $[4,3,1,2]$.

\textbf{Learning statistics rules} evaluates LLMs' ability to extract, filter, summarize, and analyze relevant information from in-context examples.  We consider four typical tasks including detecting or eliminating duplicates within the context(named duplication check and de-duplication), counting the number of elements in the context (named count \& navigation), and generating the related information about a given node in relation graph (named relation analysis). These tasks collectively aim to measure LLMs' analytical capabilities and their applications in processing and interpreting complex information structures. Figure~\ref{fig:main} shows an example of the relation analysis task that lists all the friends of "885" mentioned in the context.

\textbf{Learning list mapping} is to find diverse custom rules from given in-context examples. We use the data from "list\_functions" task in BIGBench \cite{srivastava2022beyond} to compose our numbers' list mapping task, which is learning a mapping given multi-groups of numbers list pairs. There are 250 diversity mapping rules in it with different difficulties, and one example is shown in Figure~\ref{fig:main}.

All these rule learning tasks we designed utilize n-shot examples while relying less on language abilities, commonsense knowledge, or factual knowledge.

\begin{table*}[!t]
    \centering
    \small
    \begin{tabular}{lccccccc}
        \toprule
        \multirow{3}{*}{\textbf{Model}} & \multicolumn{2}{c}{\textbf{Exact Copying}} & \multicolumn{4}{c}{\textbf{Rule Learning}} & \multirow{3}{*}{\textbf{Average}} \\
        \cline{2-3} \cline{4-7} 
        & Unstructured & Structured & \multirow{2}{*}{Format} & \multirow{2}{*}{Order} & \multirow{2}{*}{Statistics} & \multirow{2}{*}{List Mapping} &  \\
        & Text & Text & & & & & \\
        \midrule
        TinyLlama-1.1B&\underline{0.89} &0.46 &\underline{0.60} &0.46 &0.23 &0.36 &\underline{0.50} \\
        Phi-1.5&0.54 &0.25 &0.48 &0.46 &\underline{0.25} &\underline{0.39} &0.40 \\
        GPT-Neo-1.3B&0.54 &\underline{0.48} &0.48 &\underline{0.50} &0.20 &0.38 &0.43 \\
        
        \midrule
        ChatGLM3-6B&\textbf{0.90} & 0.70 &0.60 &0.54 &0.32 &0.43 &0.58 \\
        Qwen-7B&0.89 &0.45 &0.58 &0.62 &0.42 &0.48 &0.57 \\
        Baichuan2-7B&0.87 &0.44 &0.63 &0.53 &0.34 &0.49 &0.55 \\
        LLaMA-7B &0.87  &0.33  &0.68  &0.54  &0.27  &0.42  &0.52 \\
        LLaMA2-7B&0.89 &0.45 &0.61 &0.63 &0.32 &0.45 &0.56 \\
        LLaMA3-8B & 0.57 & \underline{0.87} & \underline{0.69} & \textbf{0.94} & \underline{0.61} & \underline{0.63} & \underline{0.72} \\
        Mistral-7B-v0.1 & 0.88 &0.54 &0.67 &0.78 &0.46 &0.56 & 0.65 \\
        \midrule   
        Qwen-14B&0.87 &0.48 &0.68 &0.79 &0.54 &0.55 & 0.65 \\
        Baichuan2-13B&0.80 &0.23 &\underline{0.75} &0.66 &0.40 &0.39 &0.54 \\ 
        LLaMA-13B &\underline{0.89}  &0.43  &0.63  &0.55  &0.32  &0.47  &0.55 \\
        LLaMA2-13B&\underline{0.89} &0.47 &0.60 &0.72 &0.36 &0.52 &0.59 \\
        Mixtral-8x7B & 0.88 & \underline{0.88} & 0.71 & \underline{0.86} & \underline{0.63} & \underline{0.59} & \underline{0.76}\\
        
        \midrule
        LLaMA-30B&\underline{0.89} &0.33 & 0.73 &0.77 &0.46 &0.53 &0.62 \\
        LLaMA-65B&\underline{0.89} &0.49 &0.59 & 0.85 &0.47 & 0.58 & 0.65 \\
        LLaMA2-70B& \underline{0.89} &0.43 &0.75 &0.87 &0.59 &0.58 &0.69 \\
        LLaMA3-70B& 0.64 &\textbf{0.91} &\textbf{0.90} &\textbf{0.94} &\textbf{0.87} &\textbf{0.66} &\textbf{0.82}  \\

        \bottomrule
    \end{tabular}
    \caption{Performance of LLMs with different sizes and the last column is the mean score of all aspects. A score with \underline{underline} indicates the highest score in the same group of models, while a score with \textbf{bold} indicates the highest score among all the evaluated models.}
    \label{tab:model-results}
\end{table*}

\section{Experiments}
In the experiments,
we first list the models evaluated in our work. Then, we analyze the effects of model size and pretraining tokens. Later, we give the analysis of four other factors impacting in-context learning ability. More detailed results are provided in Appendix~\ref{sec:detailed-results}.

\subsection{Settings}

\noindent\textbf{Metrics}
We have a total of 12 tasks with 2,040 testing samples. 
For almost all tasks such as string completion, dictionary search, and format conversion, we use exact match scores to evaluate the predictions with the labels. 
But for the format cloning task, we only evaluate the correctness of the format and do not consider the content. Moreover, we use postprocessing to convert models' responses for different tasks. More processing details are shown in Appendix~\ref{sec:metrics}.

\noindent\textbf{Evaluated Models}
We evaluate various open-sourced LLMs with different model sizes, such as the LLaMA series \cite{touvron2023llama, touvron2023llama2}, the Baichuan series \cite{baichuan2023baichuan2}, and the Qwen series \cite{bai2023qwen}. For example, we evaluate the LLaMA series with 7B, 13B, 34B, and 65B versions as well as base-version and chat-version. Moreover, we test the intermediate checkpoints with different pretraining stages for TinyLlama-1.1B, Baichuan-7B, and Amber-7B. The detailed description of these models can be found in Appendix~\ref{sec:models}.

\noindent\textbf{Inference Settings}
For all the models, whether base-version or chat-version, we don't use additional prompts (e.g. "User:"), and use n-shot examples in most tasks. We don't use sampling or beam search and only use the greedy decoding method. 

\begin{figure*}[!t]
    \centering
    \includegraphics[width=0.9\linewidth]{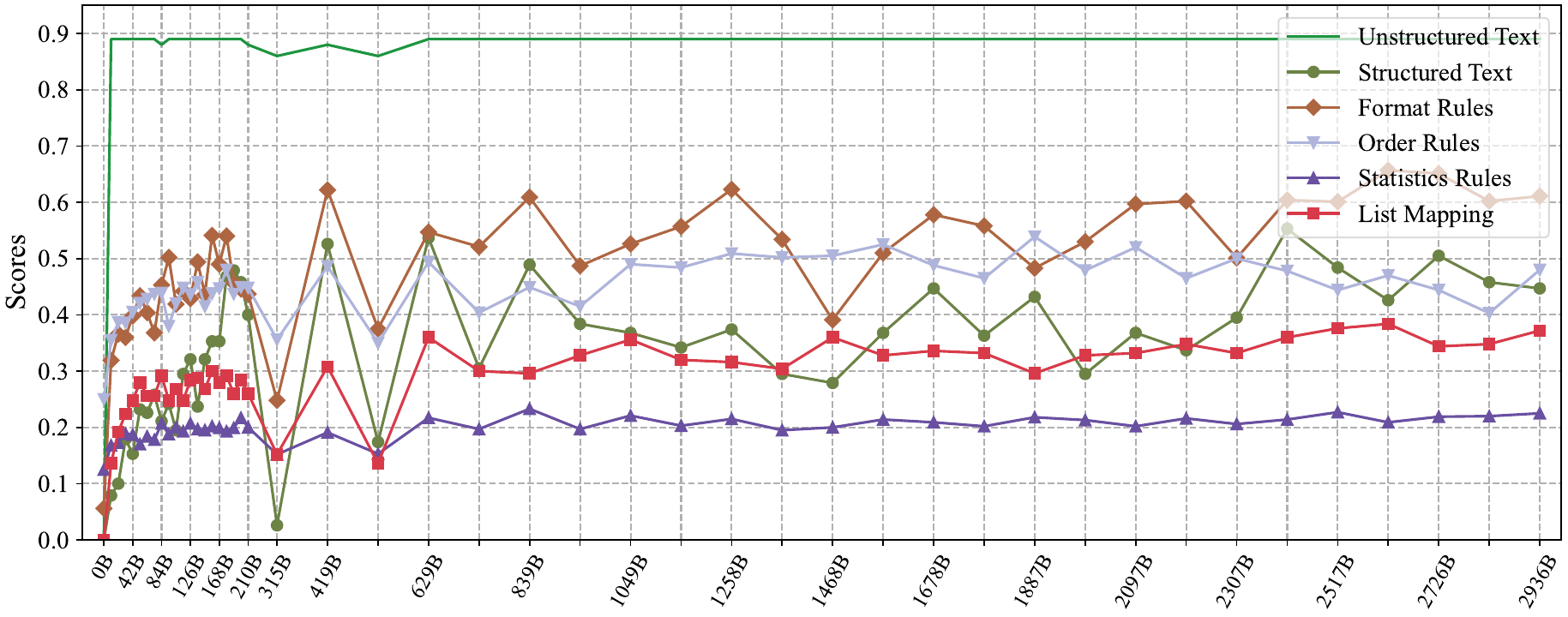}
    \caption{The scores in the pretraining stage of TinyLlama-1.1B with 3T tokens.}
    \label{fig:tinyllama-pretraining-res}
\end{figure*}

\subsection{How do Model Sizes Effect ICL Ability?}

We divided the models into three groups: small-sized models (around 1B parameters), middle-sized models (around 10B parameters), and large-sized models (bigger than 30B parameters), and conducted tests on all these models. The results are presented in Table~\ref{tab:model-results}. From the table, we can obtain the following research findings:


(1) A general trend indicates that larger models tend to exhibit superior ICL performance, observing the LLaMA series from 7B to 65B. However, the data also presents a considerable variance within models of similar sizes (e.g. LLaMA-7B, LLaMA2-7B, and LLaMA3-8B). Notably, some middle-sized models (e.g. such as Mistral-7B-v0.1 and LLaMA3-8B) demonstrate strong ICL abilities comparable to models having 5x-10x size (e.g. LLaMA-65B). This finding underscores that model size is not the sole determinant of ICL efficacy.

(2) For the exact copying ability, larger models don't have obvious advantages. Surprisingly, in the unstructured context scenario, even the small-sized models (e.g. TinyLlama-1.1B) can achieve a score of $0.89$, while the scores of LLaMA3-8B and LLaMA3-70B are only $0.57$ and $0.64$ respectively. We find that the results are mainly influenced by tokenizers, which we have a detailed discussion in Section~\ref{sec-tokenizer} and Figure~\ref{fig:bad-case-1}. In the structured context scenario, the results are also unrelated to the models' sizes. With similar model sizes, LLaMA-7B gets $0.33$ but LLaMA3-8B gets $0.87$, LLaMA2-70B gets $0.43$ but LLaMA3-70B gets $0.91$.

(3) For the rule learning ability, larger models usually have better performance than smaller ones. For challenging rules such as statistics and list mapping, the gaps between small-sized, middle-sized, and large-sized models become more obvious. The highest scores of the three groups of models are $0.25\,\text{v.s.}\,0.63, \text{v.s.}\,0.87$ in statistics rules and $0.39\,\text{v.s.}\,0.63, \text{v.s.}\,0.66$ in list mapping rules. Especially, the largest model in our evaluation (LLaMA3-70B) gets all the highest scores of the rule learning tasks. 

\begin{figure}[!t]
    \centering
    \includegraphics[width=0.89\linewidth]{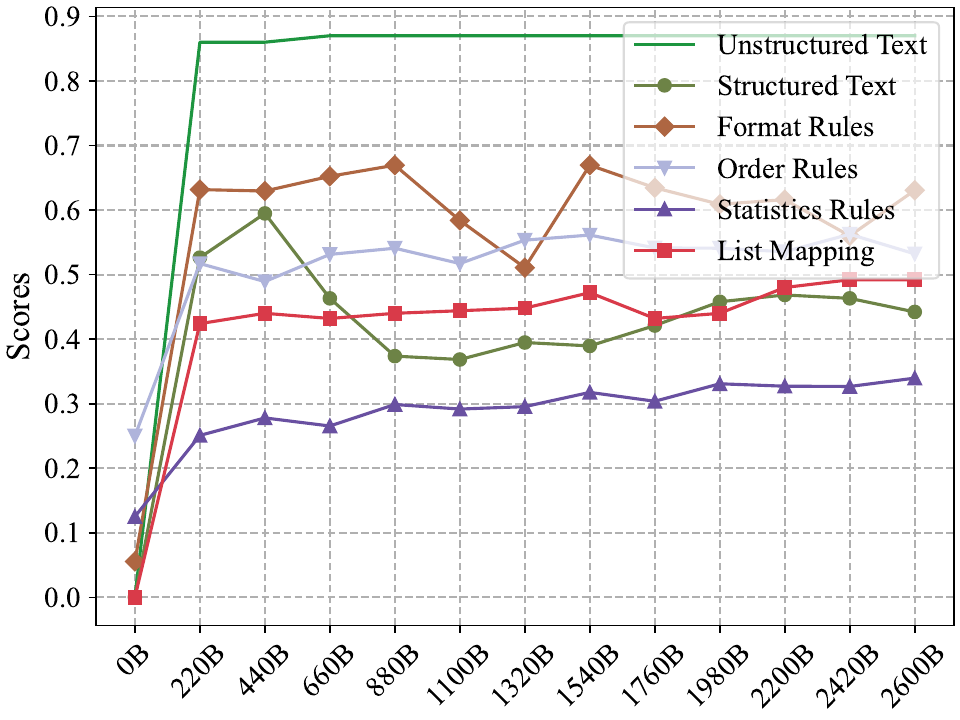}
    \caption{The scores in the pretraining stage of Baichuan2-7B with 2.6T tokens.}
    \label{fig:baichuan-pretraining-res}
\end{figure}

\begin{figure}[!t]
    \centering
    \includegraphics[width=0.89\linewidth]{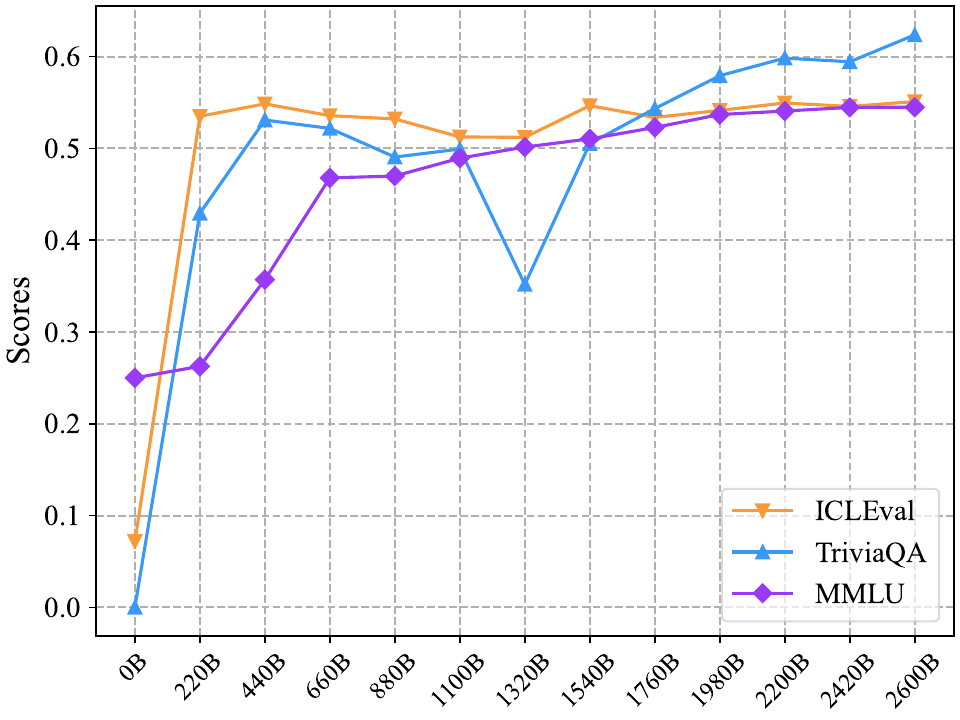}
    \caption{Baichuan2-7B's ICL ability and knowledge. The ICL ability of the model is acquired in the early stage of pretraining, while the knowledge is acquired in the whole pretraining stage.}
    \label{fig:baichuan-pretraining}
\end{figure}

\subsection{How does the ICL Ability Changing in the Pretraining Stage?}

We examine how the ICL ability evolves during the pretraining stage with the pretraining checkpoints of TinyLlama-1.1B, Baichuan2-7B, and Amber-7B. Figure~\ref{fig:tinyllama-pretraining-res} and Figure~\ref{fig:baichuan-pretraining-res} show the detailed results of TinyLlama-1.1B and Baichuan2-7B, and the results of Amber-7B are shown in Appendix~\ref{sec:pretraining-tokens}. In addition, we calculate the average scores of Baichuan2-7B and test the TriviaQA \cite{JoshiTriviaQA2017} and MMLU \cite{hendrycks2020measuring} datasets with the same checkpoints, as shown in Figure~\ref{fig:baichuan-pretraining}.

(1) The abilities of ICL exhibit rapid growth in the initial stage, before about 200B tokens. However, after this point, their growth becomes slow and eventually stops. Figure~\ref{fig:baichuan-pretraining} illustrates the results of ICLEval, TriviaQA, and MMLU, which represent the model's abilities in terms of ICL, knowledge, and processing complex questions, respectively. We can see that after 220B tokens, the ICL abilities nearly no increase, while knowledge of models continued to increase. The ability to process complex questions demonstrates little improvement before training 220B tokens but experiences a significant boost afterward. This phenomenon indicates that LLMs acquire different abilities in a sequential order, with ICL ability being relatively easy to obtain.

(2) The exact copying ability emerges in the very early stages of pretraining. As shown in Figure~\ref{fig:tinyllama-pretraining-res}, the result of copying in \textbf{unstructured text} scenarios arrives at the highest before 10B tokens and remains stable until the end of pretraining. Besides, the result of copying in \textbf{structured text} scenarios arrives at the highest score of about $0.53$ at the 419B tokens. It is unstable and fluctuates multiple times during the subsequent training stages. Compared with the unstructured text scenarios, we design various similar keys as the interference factors of prefix matching in structured text scenarios. We suppose that the distinguishing ability of similar strings is unstable in the pertaining stage.

(3) The rule learning ability increases very slowly after the initial stage, about 200B training tokens. Combined with Figure~\ref{fig:tinyllama-pretraining-res} and Figure~\ref{fig:baichuan-pretraining-res}, we can find that the results of learning format rules are unstable in the pertaining stage, while the results of the other three aspects rules are more stable. We speculate that format learning is more easily interference by the LLMs' inherent preferences, and these inherent preferences continuously change during the training stage. Inherent preferences will be discussed in Section~\ref{inherent-preference}.

In particular, there is a significant increase in statistics for Baichuan-7B but only a slight improvement in TinyLlama-1.1B. We guess that maybe models only learn to predict the next token based on a small number of previous tokens in the early stages of pretraining. While, during the later stage of pretraining, models learn to use more tokens to predict the next token more accurately. Small-sized models may not have enough attention points capacity for later pretraining, which will be discussed in detail in Section~\ref{attention-points-capacity}.

\subsection{Case Study.}

We find some interesting phenomenons in our evaluation results and regard these phenomena to four aspects: distinguishing ability, inherent preferences, attention points capacity, and tokenizer. We also analyze these phenomena with some bad cases. Due to the limitation of pages, we put the bad cases on the Appendix~\ref{sec:bad case}. 

\begin{figure}[!t]
    \centering
    \includegraphics[width=0.9\linewidth]{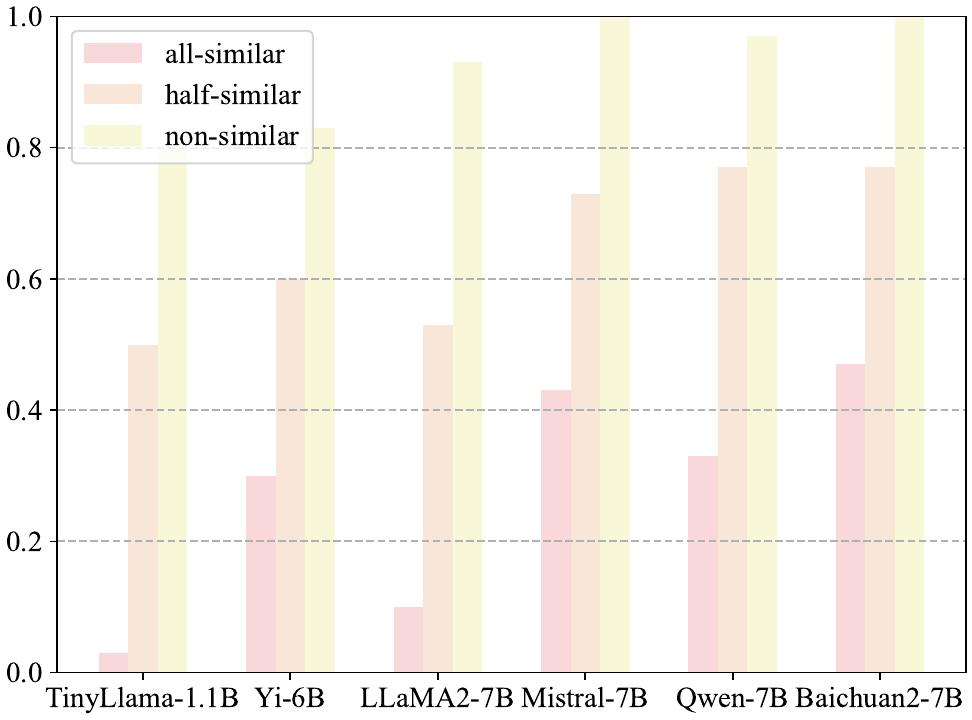}
    \caption{Performance changes when there are more similar strings in the in-context examples.}
    
    \label{fig:similar-copying}
\end{figure}

\begin{figure*}[!]
    \centering
    \includegraphics[width=0.85\linewidth]{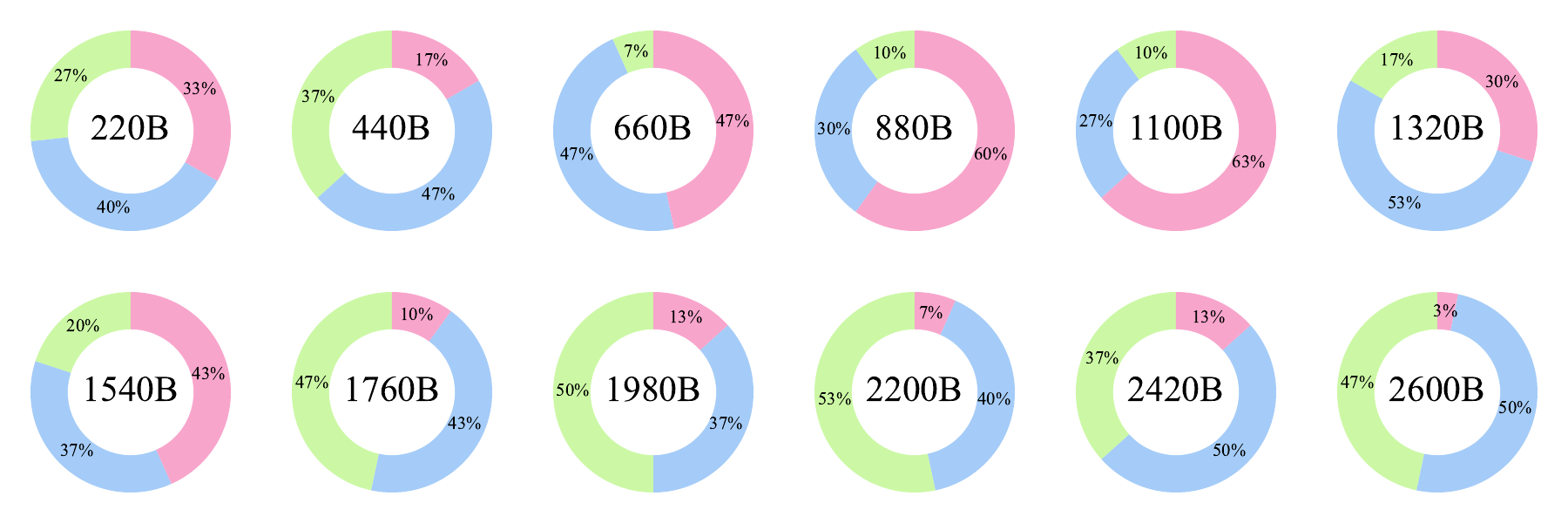}
    \caption{The proportion of three answer types in the pretraining stage of Baichuan2-7B. \textcolor{darkgreen}{Green} means copying right value, \textcolor{skyblue}{blue} means copying incorrect values, and \textcolor{darkpink}{pink} means genrating random values. We find the model's choices are continuously changing in the pretraining stage.} 
    \label{fig:baichaun-distribution}
\end{figure*}

\paragraph{Distinguishing Ability}

Similar strings can cause interference for humans as well as for LLMs. As depicted in Figure~\ref{fig:similar-copying}, in the dictionary search task, the accuracy of the same model is adversely affected when there are numerous similar keys in the dictionary.  
More similar keys make the scores drop more, while models with the stronger distinguishing ability (e.g. Baichuan2-7B) drop less. The results suggest that the excessive presence of similar strings makes the model chaotic, causing it to struggle to distinguish between different keys. This phenomenon indicates that models with weak distinguishing ability may make mistakes when extracting information in complex in-context environments.

We looked into the reasons for errors in certain cases and found two types. First, the model sometimes selects the incorrect value from the dictionary. Second, it occasionally generates a random string that is not present in the dictionary. We further analyze the proportion of right results and the two error types in the pretraining stage of Baichuan2-7B, as shown in Figure~\ref{fig:baichaun-distribution}. We find the model's choices of the two types are continuously changing. That indicates the distinguishing ability is unstable in the pretraining stage.

\paragraph{Inherent Preferences}
\label{inherent-preference}

\begin{figure}[!]
    \centering
    \includegraphics[width=1\linewidth]{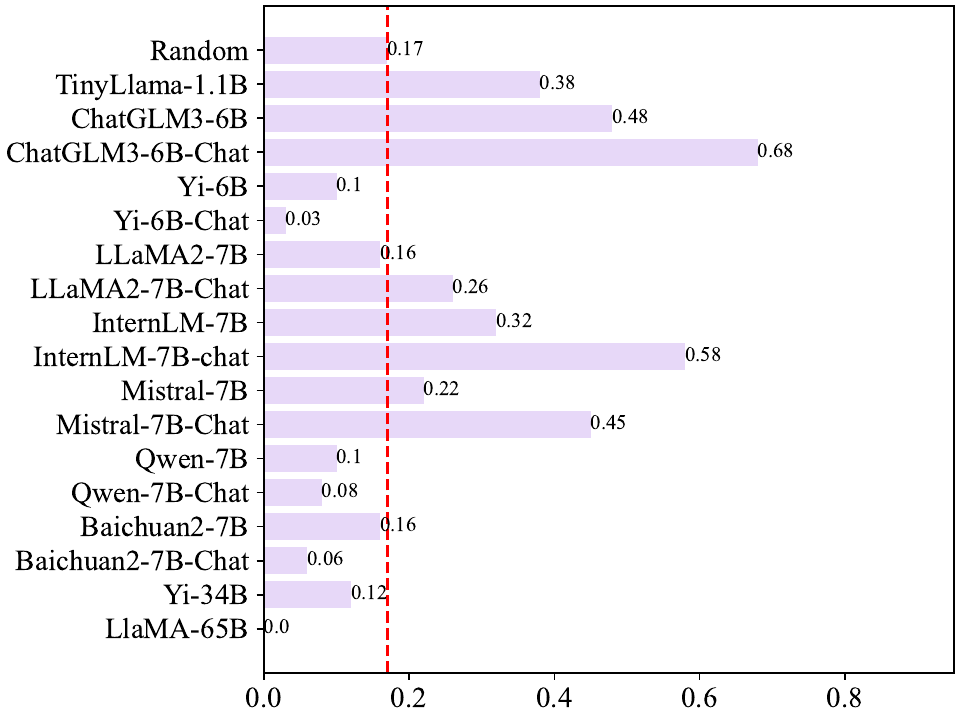}
    \caption{Performance of format check task.}
    \label{fig:preference-format}
\end{figure}

We observe that some models exhibit unusually weak performance in tasks such as format check or format cloning. Figure~\ref{fig:preference-format} shows the performance of the format check task. This task is a classification task with six different labels: "JSONL", "CSV", "Tuple", "YAML", "Table", and "XML". From the figure, we can find that ChatGLM3-6B-Chat can get a score approaching 0.7, while lots of models are lower than the random scores. To our surprise, LLaMA-65B gets 0.0 in this task. 

We further give a deeper analysis of the bad cases, and we find that most models tend to respond with "JSON" as their prediction, but we even don't have this label. We suppose that such heavy inherent preferences of these models may come from their pretraining data distribution, making the model cannot adhere to the pre-defined formats or rules presented in the in-context examples. 

Furthermore, we find the chat version of some models such as ChatGLM3-6B, InternLM-7B, and Mistral-7B have obvious improvements to their base version in this task. We think this might be due to the instruction learning process can reduce the impact of the models' inherent preferences to some extent.

\paragraph{Attention Points Capacity}
\label{attention-points-capacity}

In the count \& navigation task, the "1-dim" setting requires models to analyze and count only two elements ("up" and "down"), whereas the "2-dim" setting involves analyzing and counting four elements ("up", "down", "right" and "left"). We have observed that the scores for the "1-dim" setting are significantly higher than those for the "2-dim" setting, as illustrated in Table~\ref{tab:count-naviagtion-forms}. 

\begin{figure}[!t]
    \centering
    \includegraphics[width=1.0\linewidth]{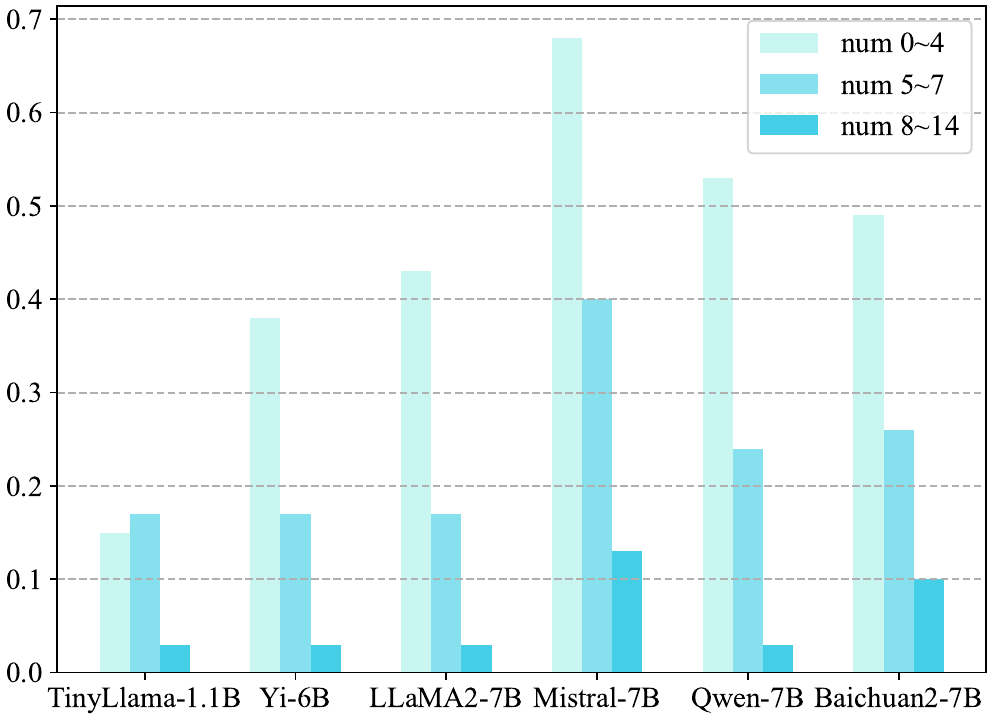}
    \caption{The scores of count \& navigation task. We split the task by the statistics elements' number. More elements indicate more need for attention points.}
    \label{fig:count-navigation}
\end{figure}

We believe this phenomenon occurs because models face difficulties in effectively utilizing a larger number of tokens within the context to predict the next token. In other words, for models to accurately predict results, they need to pay attention to multiple points within the context. And we call this Attention Points Capacity. However, many models tend to rely on only a few tokens for predicting the next token, thereby failing to incorporate all the available information simultaneously. To validate our hypothesis, we conducted a further analysis of the relationship between the number of elements and accuracy, as depicted in Figure~\ref{fig:count-navigation}. The figure demonstrates a significant decrease in accuracy as the number of elements increases.

\begin{table}[!t]
    \centering
    \small
    \begin{tabular}{lcccc}
        \toprule
        \multirow{2}{*}{\textbf{Model}} & \multicolumn{2}{c}{\textbf{Count}} & \multicolumn{2}{c}{\textbf{Navigation}} \\
        & 1-dim & 2-dim & 1-dim & 2-dim \\
        \midrule
        TinyLlama-1.1B & 0.17 & 0.03 & 0.27 & 0.03 \\
        Yi-6B& 0.40 & 0.00 & 0.43 & 0.03 \\
        LLaMA2-7B& 0.47 & 0.07 & 0.27 & 0.13 \\
        Mistral-7B-v0.1& 0.77 & 0.43 & 0.47 & 0.10 \\
        Qwen-7B& 0.60 & 0.10 & 0.53 & 0.10 \\
        Baichuan2-7B& 0.77 & 0.07 & 0.33 & 0.07 \\
        \bottomrule
    \end{tabular}
    \caption{The scores of count \& navigation task. We split the tasks by task types.}
    \label{tab:count-naviagtion-forms}
\end{table}

\paragraph{Tokenizer}
\label{sec-tokenizer}
In our evaluation, we observe that the tokenizer has a significant impact on our results and can affect our adjustment of task difficulty. From a human perspective, we can perceive text at the character level, word level, or sentence level, allowing us to easily distinguish individual numbers, letters, words, and sentences. However, language models process inputs at the token level, which presents a different viewpoint compared to ours. 

\begin{figure}[!t]
    \centering
    \includegraphics[width=1\linewidth]{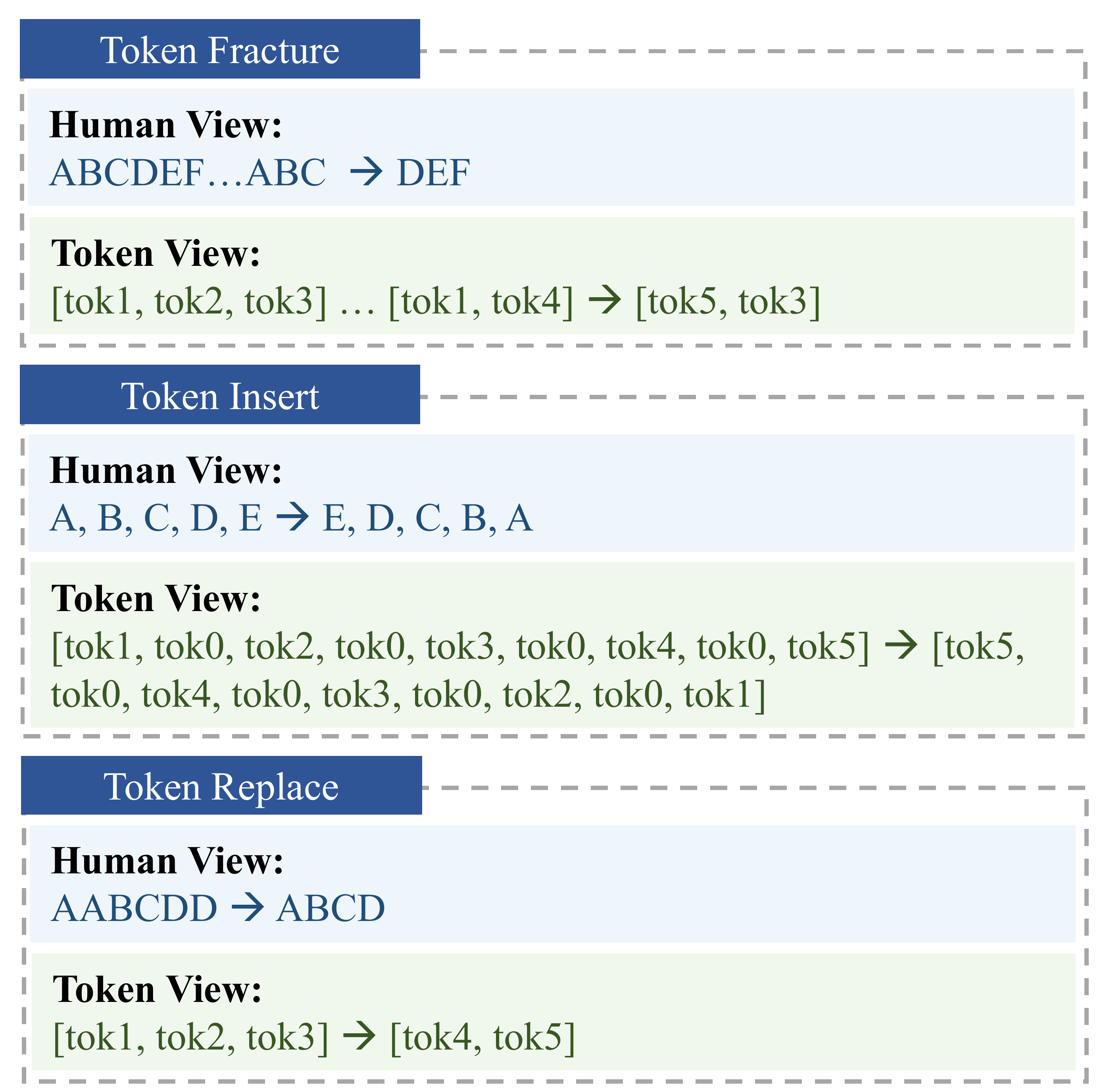}
    \caption{The three tokenization phenomena in which token views are different from human views.}
    \label{fig:tokenizer}
\end{figure}

In Figure~\ref{fig:tokenizer}, we illustrate three types of bad cases caused by tokenization that we find can lead to confusion for LLMs. "Token fracture" occurs when a continuous sequence is split but the encoded tokens of the segmented fragments do not match those of the original sequence; "Token insert" refers to the fact that the special symbols that we can easily overlook also may require 1-3 tokens for encoding; "Token replace" indicates that several consecutive characters can be encoded as a single token, resulting in different encoded tokens for the reversed string compared to the original one.

\section{Related Work}
\paragraph{In-Context Learning} The generalization ability of language models has always been a goal pursued by researchers, and the discovery of in-context learning ability is a milestone in achieving this goal. It was discovered in GPT-3 \citep{brown2020language} that the model can implement custom outputs based on instructions and examples. In-context learning can enable models to adapt to new tasks and even learn new knowledge and rules without updating model parameters. The phenomenon of in-context learning has attracted widespread attention from researchers to explore \citep{lu2021fantastically, nie2022improving, ye2022complementary, min2022rethinking, liu2023lost}, explain \citep{xie2021explanation, akyurek2022learning, han2023context, li2023transformers}, enhance \citep{min2021metaicl, chen2021meta, yang2023iterative}, and apply \citep{dua2022successive, shridhar2022distilling, wu2023openicl} it. Chain-of-thought (CoT) reasoning is special in-context learning, divided into two modes: few-shot CoT \citep{wei2022chain} and zero-shot CoT \citep{kojima2022large}. Few-shot prompts is an important part of CoT, and many studies \citep{wang2022self, zhou2022least, fu2022complexity, lyu2023faithful, shum2023automatic, diao2023active} focus on selecting better examples to promote the application of CoT. we propose an \ourdata benchmark for measuring in-context learning abilities. To the best of our knowledge, it's the first work to estimate ICL abilities comprehensively.

\paragraph{Model Evaluation} 
After the emergence of the large model, to comprehensively understand the boundaries and behaviors of its abilities, researchers tested the model from multiple aspects such as language ability, knowledge, professional ability, theory of mind, and security. The testing of model language abilities can utilize various traditional NLP task datasets, including language comprehension \citep{qin2023chatgpt}, language generation \citep{qin2023chatgpt}, and multilingualism \citep{lai2023chatgpt}. The testing of model knowledge includes common sense knowledge \citep{clark2018think, bisk2020piqa, zellers2019hellaswag, mihaylov2018can}, factual knowledge \citep{kwiatkowski2019natural, lin2021truthfulqa, berant2013semantic}, technical knowledge \citep{yuan2023well}, etc. The Achievement test for model specialty includes mathematics \citep{cobbe2021training, ling2017program, hendrycks2021measuring}, coding \citep{austin2021program, chen2021evaluating}, medicine \citep{thirunavukarasu2023trialling}, etc.In \citep{bubeck2023sparks}, theory of mind is considered to test GPT-4\citep{gpt4}. The test of model security includes stability, ethics, biases, and hallucination. Among them, many benchmarks can perform relatively comprehensive evaluation on the models, such as HELM\citep{liang2022holistic}, MMLU \citep{hendrycks2020measuring}, C\_EVAL\citep{huang2023c}, AGIEval\citep{zhong2023agieval},  BIGBench\citep{srivastava2022beyond}, etc. However, most of the evaluations here need to involve a variety of different basic capabilities, and the quality of the test results is affected by a variety of factors. Our \ourdata benchmark decouples the evaluation of the in-context learning ability from other abilities. 

\section{Conclusion}
In this work, we introduce an \ourdata benchmark for measuring the in-context learning ability of LLMs.
We classify it into copying ability and learning ability, and design 12 evaluation tasks. We find that model size is an important but not the sole determinant of ICL ability. Also, we observe that ICL abilities, particularly copying ability have a quick increase in the very early pretraining stage. Furthermore, we analyze some bad cases and discover that the ICL abilities are influenced by distinguishing ability, inherent preferences, attention points capacity, and tokenizer. That indicates that we need to consider these factors if we would like to train a model with strong in-context ability.

\section*{Limitations}
When designing the \ourdata evaluation, we minimize the need for other abilities of the model as much as possible. However, we still can’t guarantee that the testing tasks will not be affected by the language abilities and internal knowledge of the model itself. For example, the inherent preferences we discussed in Section~\ref{inherent-preference} might make LLMs ignore the context.

\section*{Acknowledgements}
We sincerely thank all the anonymous reviewers and (S)ACs for their constructive comments and helpful suggestions. This work was supported by The National Natural Science Foundation of China (No.\ 62376273).

\bibliography{custom}

\begin{thebibliography}{62}
\expandafter\ifx\csname natexlab\endcsname\relax\def\natexlab#1{#1}\fi

\bibitem[{Aky{\"u}rek et~al.(2022)Aky{\"u}rek, Schuurmans, Andreas, Ma, and Zhou}]{akyurek2022learning}
Ekin Aky{\"u}rek, Dale Schuurmans, Jacob Andreas, Tengyu Ma, and Denny Zhou. 2022.
\newblock What learning algorithm is in-context learning? investigations with linear models.
\newblock \emph{arXiv preprint arXiv:2211.15661}.

\bibitem[{Austin et~al.(2021)Austin, Odena, Nye, Bosma, Michalewski, Dohan, Jiang, Cai, Terry, Le et~al.}]{austin2021program}
Jacob Austin, Augustus Odena, Maxwell Nye, Maarten Bosma, Henryk Michalewski, David Dohan, Ellen Jiang, Carrie Cai, Michael Terry, Quoc Le, et~al. 2021.
\newblock Program synthesis with large language models.
\newblock \emph{arXiv preprint arXiv:2108.07732}.

\bibitem[{Bai et~al.(2023)Bai, Bai, Chu, Cui, Dang, Deng, Fan, Ge, Han, Huang et~al.}]{bai2023qwen}
Jinze Bai, Shuai Bai, Yunfei Chu, Zeyu Cui, Kai Dang, Xiaodong Deng, Yang Fan, Wenbin Ge, Yu~Han, Fei Huang, et~al. 2023.
\newblock Qwen technical report.
\newblock \emph{arXiv preprint arXiv:2309.16609}.

\bibitem[{Baichuan(2023)}]{baichuan2023baichuan2}
Baichuan. 2023.
\newblock \href {https://arxiv.org/abs/2309.10305} {Baichuan 2: Open large-scale language models}.
\newblock \emph{arXiv preprint arXiv:2309.10305}.

\bibitem[{Berant et~al.(2013)Berant, Chou, Frostig, and Liang}]{berant2013semantic}
Jonathan Berant, Andrew Chou, Roy Frostig, and Percy Liang. 2013.
\newblock Semantic parsing on freebase from question-answer pairs.
\newblock In \emph{Proceedings of the 2013 conference on empirical methods in natural language processing}, pages 1533--1544.

\bibitem[{Bisk et~al.(2020)Bisk, Zellers, Gao, Choi et~al.}]{bisk2020piqa}
Yonatan Bisk, Rowan Zellers, Jianfeng Gao, Yejin Choi, et~al. 2020.
\newblock Piqa: Reasoning about physical commonsense in natural language.
\newblock In \emph{Proceedings of the AAAI conference on artificial intelligence}, volume~34, pages 7432--7439.

\bibitem[{Brown et~al.(2020)Brown, Mann, Ryder, Subbiah, Kaplan, Dhariwal, Neelakantan, Shyam, Sastry, Askell et~al.}]{brown2020language}
Tom Brown, Benjamin Mann, Nick Ryder, Melanie Subbiah, Jared~D Kaplan, Prafulla Dhariwal, Arvind Neelakantan, Pranav Shyam, Girish Sastry, Amanda Askell, et~al. 2020.
\newblock Language models are few-shot learners.
\newblock \emph{Advances in neural information processing systems}, 33:1877--1901.

\bibitem[{Bubeck et~al.(2023)Bubeck, Chandrasekaran, Eldan, Gehrke, Horvitz, Kamar, Lee, Lee, Li, Lundberg et~al.}]{bubeck2023sparks}
S{\'e}bastien Bubeck, Varun Chandrasekaran, Ronen Eldan, Johannes Gehrke, Eric Horvitz, Ece Kamar, Peter Lee, Yin~Tat Lee, Yuanzhi Li, Scott Lundberg, et~al. 2023.
\newblock Sparks of artificial general intelligence: Early experiments with gpt-4.
\newblock \emph{arXiv preprint arXiv:2303.12712}.

\bibitem[{Chen et~al.(2021{\natexlab{a}})Chen, Tworek, Jun, Yuan, Pinto, Kaplan, Edwards, Burda, Joseph, Brockman et~al.}]{chen2021evaluating}
Mark Chen, Jerry Tworek, Heewoo Jun, Qiming Yuan, Henrique Ponde de~Oliveira Pinto, Jared Kaplan, Harri Edwards, Yuri Burda, Nicholas Joseph, Greg Brockman, et~al. 2021{\natexlab{a}}.
\newblock Evaluating large language models trained on code.
\newblock \emph{arXiv preprint arXiv:2107.03374}.

\bibitem[{Chen et~al.(2021{\natexlab{b}})Chen, Zhong, Zha, Karypis, and He}]{chen2021meta}
Yanda Chen, Ruiqi Zhong, Sheng Zha, George Karypis, and He~He. 2021{\natexlab{b}}.
\newblock Meta-learning via language model in-context tuning.
\newblock \emph{arXiv preprint arXiv:2110.07814}.

\bibitem[{Clark et~al.(2018)Clark, Cowhey, Etzioni, Khot, Sabharwal, Schoenick, and Tafjord}]{clark2018think}
Peter Clark, Isaac Cowhey, Oren Etzioni, Tushar Khot, Ashish Sabharwal, Carissa Schoenick, and Oyvind Tafjord. 2018.
\newblock Think you have solved question answering? try arc, the ai2 reasoning challenge.
\newblock \emph{arXiv preprint arXiv:1803.05457}.

\bibitem[{Cobbe et~al.(2021)Cobbe, Kosaraju, Bavarian, Chen, Jun, Kaiser, Plappert, Tworek, Hilton, Nakano et~al.}]{cobbe2021training}
Karl Cobbe, Vineet Kosaraju, Mohammad Bavarian, Mark Chen, Heewoo Jun, Lukasz Kaiser, Matthias Plappert, Jerry Tworek, Jacob Hilton, Reiichiro Nakano, et~al. 2021.
\newblock Training verifiers to solve math word problems.
\newblock \emph{arXiv preprint arXiv:2110.14168}.

\bibitem[{Diao et~al.(2023)Diao, Wang, Lin, and Zhang}]{diao2023active}
Shizhe Diao, Pengcheng Wang, Yong Lin, and Tong Zhang. 2023.
\newblock Active prompting with chain-of-thought for large language models.
\newblock \emph{arXiv preprint arXiv:2302.12246}.

\bibitem[{Dua et~al.(2022)Dua, Gupta, Singh, and Gardner}]{dua2022successive}
Dheeru Dua, Shivanshu Gupta, Sameer Singh, and Matt Gardner. 2022.
\newblock Successive prompting for decomposing complex questions.
\newblock \emph{arXiv preprint arXiv:2212.04092}.

\bibitem[{Fu et~al.(2022)Fu, Peng, Sabharwal, Clark, and Khot}]{fu2022complexity}
Yao Fu, Hao Peng, Ashish Sabharwal, Peter Clark, and Tushar Khot. 2022.
\newblock Complexity-based prompting for multi-step reasoning.
\newblock \emph{arXiv preprint arXiv:2210.00720}.

\bibitem[{Han et~al.(2023)Han, Wang, Zhao, and Ji}]{han2023context}
Chi Han, Ziqi Wang, Han Zhao, and Heng Ji. 2023.
\newblock In-context learning of large language models explained as kernel regression.
\newblock \emph{arXiv preprint arXiv:2305.12766}.

\bibitem[{Hendrycks et~al.(2020)Hendrycks, Burns, Basart, Zou, Mazeika, Song, and Steinhardt}]{hendrycks2020measuring}
Dan Hendrycks, Collin Burns, Steven Basart, Andy Zou, Mantas Mazeika, Dawn Song, and Jacob Steinhardt. 2020.
\newblock Measuring massive multitask language understanding.
\newblock \emph{arXiv preprint arXiv:2009.03300}.

\bibitem[{Hendrycks et~al.(2021)Hendrycks, Burns, Kadavath, Arora, Basart, Tang, Song, and Steinhardt}]{hendrycks2021measuring}
Dan Hendrycks, Collin Burns, Saurav Kadavath, Akul Arora, Steven Basart, Eric Tang, Dawn Song, and Jacob Steinhardt. 2021.
\newblock Measuring mathematical problem solving with the math dataset.
\newblock \emph{arXiv preprint arXiv:2103.03874}.

\bibitem[{Huang et~al.(2023)Huang, Bai, Zhu, Zhang, Zhang, Su, Liu, Lv, Zhang, Lei et~al.}]{huang2023c}
Yuzhen Huang, Yuzhuo Bai, Zhihao Zhu, Junlei Zhang, Jinghan Zhang, Tangjun Su, Junteng Liu, Chuancheng Lv, Yikai Zhang, Jiayi Lei, et~al. 2023.
\newblock C-eval: A multi-level multi-discipline chinese evaluation suite for foundation models.
\newblock \emph{arXiv preprint arXiv:2305.08322}.

\bibitem[{Joshi et~al.(2017)Joshi, Choi, Weld, and Zettlemoyer}]{JoshiTriviaQA2017}
Mandar Joshi, Eunsol Choi, Daniel~S. Weld, and Luke Zettlemoyer. 2017.
\newblock Triviaqa: A large scale distantly supervised challenge dataset for reading comprehension.
\newblock In \emph{Proceedings of the 55th Annual Meeting of the Association for Computational Linguistics}, Vancouver, Canada. Association for Computational Linguistics.

\bibitem[{Kojima et~al.(2022)Kojima, Gu, Reid, Matsuo, and Iwasawa}]{kojima2022large}
Takeshi Kojima, Shixiang~Shane Gu, Machel Reid, Yutaka Matsuo, and Yusuke Iwasawa. 2022.
\newblock Large language models are zero-shot reasoners.
\newblock \emph{Advances in neural information processing systems}, 35:22199--22213.

\bibitem[{Kwiatkowski et~al.(2019)Kwiatkowski, Palomaki, Redfield, Collins, Parikh, Alberti, Epstein, Polosukhin, Devlin, Lee et~al.}]{kwiatkowski2019natural}
Tom Kwiatkowski, Jennimaria Palomaki, Olivia Redfield, Michael Collins, Ankur Parikh, Chris Alberti, Danielle Epstein, Illia Polosukhin, Jacob Devlin, Kenton Lee, et~al. 2019.
\newblock Natural questions: a benchmark for question answering research.
\newblock \emph{Transactions of the Association for Computational Linguistics}, 7:453--466.

\bibitem[{Lai et~al.(2017)Lai, Xie, Liu, Yang, and Hovy}]{lai2017race}
Guokun Lai, Qizhe Xie, Hanxiao Liu, Yiming Yang, and Eduard Hovy. 2017.
\newblock Race: Large-scale reading comprehension dataset from examinations.
\newblock \emph{arXiv preprint arXiv:1704.04683}.

\bibitem[{Lai et~al.(2023)Lai, Ngo, Veyseh, Man, Dernoncourt, Bui, and Nguyen}]{lai2023chatgpt}
Viet~Dac Lai, Nghia~Trung Ngo, Amir Pouran~Ben Veyseh, Hieu Man, Franck Dernoncourt, Trung Bui, and Thien~Huu Nguyen. 2023.
\newblock Chatgpt beyond english: Towards a comprehensive evaluation of large language models in multilingual learning.
\newblock \emph{arXiv preprint arXiv:2304.05613}.

\bibitem[{Li et~al.(2023)Li, Ildiz, Papailiopoulos, and Oymak}]{li2023transformers}
Yingcong Li, Muhammed~Emrullah Ildiz, Dimitris Papailiopoulos, and Samet Oymak. 2023.
\newblock Transformers as algorithms: Generalization and stability in in-context learning.

\bibitem[{Liang et~al.(2022)Liang, Bommasani, Lee, Tsipras, Soylu, Yasunaga, Zhang, Narayanan, Wu, Kumar et~al.}]{liang2022holistic}
Percy Liang, Rishi Bommasani, Tony Lee, Dimitris Tsipras, Dilara Soylu, Michihiro Yasunaga, Yian Zhang, Deepak Narayanan, Yuhuai Wu, Ananya Kumar, et~al. 2022.
\newblock Holistic evaluation of language models.
\newblock \emph{arXiv preprint arXiv:2211.09110}.

\bibitem[{Lin et~al.(2021)Lin, Hilton, and Evans}]{lin2021truthfulqa}
Stephanie Lin, Jacob Hilton, and Owain Evans. 2021.
\newblock Truthfulqa: Measuring how models mimic human falsehoods.
\newblock \emph{arXiv preprint arXiv:2109.07958}.

\bibitem[{Ling et~al.(2017)Ling, Yogatama, Dyer, and Blunsom}]{ling2017program}
Wang Ling, Dani Yogatama, Chris Dyer, and Phil Blunsom. 2017.
\newblock Program induction by rationale generation: Learning to solve and explain algebraic word problems.
\newblock \emph{arXiv preprint arXiv:1705.04146}.

\bibitem[{Liu et~al.(2023{\natexlab{a}})Liu, Lin, Hewitt, Paranjape, Bevilacqua, Petroni, and Liang}]{liu2023lost}
Nelson~F Liu, Kevin Lin, John Hewitt, Ashwin Paranjape, Michele Bevilacqua, Fabio Petroni, and Percy Liang. 2023{\natexlab{a}}.
\newblock Lost in the middle: How language models use long contexts.
\newblock \emph{arXiv preprint arXiv:2307.03172}.

\bibitem[{Liu et~al.(2023{\natexlab{b}})Liu, Qiao, Neiswanger, Wang, Tan, Tao, Li, Wang, Sun, Pangarkar, Fan, Gu, Miller, Zhuang, He, Li, Koto, Tang, Ranjan, Shen, Ren, Iriondo, Mu, Hu, Schulze, Nakov, Baldwin, and Xing}]{liu2023llm360}
Zhengzhong Liu, Aurick Qiao, Willie Neiswanger, Hongyi Wang, Bowen Tan, Tianhua Tao, Junbo Li, Yuqi Wang, Suqi Sun, Omkar Pangarkar, Richard Fan, Yi~Gu, Victor Miller, Yonghao Zhuang, Guowei He, Haonan Li, Fajri Koto, Liping Tang, Nikhil Ranjan, Zhiqiang Shen, Xuguang Ren, Roberto Iriondo, Cun Mu, Zhiting Hu, Mark Schulze, Preslav Nakov, Tim Baldwin, and Eric~P. Xing. 2023{\natexlab{b}}.
\newblock \href {http://arxiv.org/abs/2312.06550} {Llm360: Towards fully transparent open-source llms}.

\bibitem[{Lu et~al.(2021)Lu, Bartolo, Moore, Riedel, and Stenetorp}]{lu2021fantastically}
Yao Lu, Max Bartolo, Alastair Moore, Sebastian Riedel, and Pontus Stenetorp. 2021.
\newblock Fantastically ordered prompts and where to find them: Overcoming few-shot prompt order sensitivity.
\newblock \emph{arXiv preprint arXiv:2104.08786}.

\bibitem[{Lyu et~al.(2023)Lyu, Havaldar, Stein, Zhang, Rao, Wong, Apidianaki, and Callison-Burch}]{lyu2023faithful}
Qing Lyu, Shreya Havaldar, Adam Stein, Li~Zhang, Delip Rao, Eric Wong, Marianna Apidianaki, and Chris Callison-Burch. 2023.
\newblock Faithful chain-of-thought reasoning.
\newblock \emph{arXiv preprint arXiv:2301.13379}.

\bibitem[{Mihaylov et~al.(2018)Mihaylov, Clark, Khot, and Sabharwal}]{mihaylov2018can}
Todor Mihaylov, Peter Clark, Tushar Khot, and Ashish Sabharwal. 2018.
\newblock Can a suit of armor conduct electricity? a new dataset for open book question answering.
\newblock \emph{arXiv preprint arXiv:1809.02789}.

\bibitem[{Min et~al.(2021)Min, Lewis, Zettlemoyer, and Hajishirzi}]{min2021metaicl}
Sewon Min, Mike Lewis, Luke Zettlemoyer, and Hannaneh Hajishirzi. 2021.
\newblock Metaicl: Learning to learn in context.
\newblock \emph{arXiv preprint arXiv:2110.15943}.

\bibitem[{Min et~al.(2022)Min, Lyu, Holtzman, Artetxe, Lewis, Hajishirzi, and Zettlemoyer}]{min2022rethinking}
Sewon Min, Xinxi Lyu, Ari Holtzman, Mikel Artetxe, Mike Lewis, Hannaneh Hajishirzi, and Luke Zettlemoyer. 2022.
\newblock Rethinking the role of demonstrations: What makes in-context learning work?
\newblock \emph{arXiv preprint arXiv:2202.12837}.

\bibitem[{Nie et~al.(2022)Nie, Chen, Zhang, and Cheng}]{nie2022improving}
Feng Nie, Meixi Chen, Zhirui Zhang, and Xu~Cheng. 2022.
\newblock Improving few-shot performance of language models via nearest neighbor calibration.
\newblock \emph{arXiv preprint arXiv:2212.02216}.

\bibitem[{Olsson et~al.(2022)Olsson, Elhage, Nanda, Joseph, DasSarma, Henighan, Mann, Askell, Bai, Chen et~al.}]{olsson2022context}
Catherine Olsson, Nelson Elhage, Neel Nanda, Nicholas Joseph, Nova DasSarma, Tom Henighan, Ben Mann, Amanda Askell, Yuntao Bai, Anna Chen, et~al. 2022.
\newblock In-context learning and induction heads.
\newblock \emph{arXiv preprint arXiv:2209.11895}.

\bibitem[{OpenAI(2023)}]{gpt4}
OpenAI. 2023.
\newblock Gpt-4 technical report.

\bibitem[{Peiyuan~Zhang and Lu(2023)}]{tinyllama}
Tianduo~Wang Peiyuan~Zhang, Guangtao~Zeng and Wei Lu. 2023.
\newblock \href {https://github.com/jzhang38/TinyLlama} {Tinyllama}.

\bibitem[{Qin et~al.(2023{\natexlab{a}})Qin, Zhang, Zhang, Chen, Yasunaga, and Yang}]{qin2023chatgpt}
Chengwei Qin, Aston Zhang, Zhuosheng Zhang, Jiaao Chen, Michihiro Yasunaga, and Diyi Yang. 2023{\natexlab{a}}.
\newblock Is chatgpt a general-purpose natural language processing task solver?
\newblock \emph{arXiv preprint arXiv:2302.06476}.

\bibitem[{Qin et~al.(2023{\natexlab{b}})Qin, Hu, Lin, Chen, Ding, Cui, Zeng, Huang, Xiao, Han et~al.}]{qin2023tool}
Yujia Qin, Shengding Hu, Yankai Lin, Weize Chen, Ning Ding, Ganqu Cui, Zheni Zeng, Yufei Huang, Chaojun Xiao, Chi Han, et~al. 2023{\natexlab{b}}.
\newblock Tool learning with foundation models.
\newblock \emph{arXiv preprint arXiv:2304.08354}.

\bibitem[{Rajpurkar et~al.(2018)Rajpurkar, Jia, and Liang}]{rajpurkar2018know}
Pranav Rajpurkar, Robin Jia, and Percy Liang. 2018.
\newblock Know what you don't know: Unanswerable questions for squad.
\newblock \emph{arXiv preprint arXiv:1806.03822}.

\bibitem[{Reddy et~al.(2019)Reddy, Chen, and Manning}]{reddy2019coqa}
Siva Reddy, Danqi Chen, and Christopher~D Manning. 2019.
\newblock Coqa: A conversational question answering challenge.
\newblock \emph{Transactions of the Association for Computational Linguistics}, 7:249--266.

\bibitem[{Shridhar et~al.(2022)Shridhar, Stolfo, and Sachan}]{shridhar2022distilling}
Kumar Shridhar, Alessandro Stolfo, and Mrinmaya Sachan. 2022.
\newblock Distilling multi-step reasoning capabilities of large language models into smaller models via semantic decompositions.
\newblock \emph{arXiv preprint arXiv:2212.00193}.

\bibitem[{Shum et~al.(2023)Shum, Diao, and Zhang}]{shum2023automatic}
KaShun Shum, Shizhe Diao, and Tong Zhang. 2023.
\newblock Automatic prompt augmentation and selection with chain-of-thought from labeled data.
\newblock \emph{arXiv preprint arXiv:2302.12822}.

\bibitem[{Srivastava et~al.(2022)Srivastava, Rastogi, Rao, Shoeb, Abid, Fisch, Brown, Santoro, Gupta, Garriga-Alonso et~al.}]{srivastava2022beyond}
Aarohi Srivastava, Abhinav Rastogi, Abhishek Rao, Abu Awal~Md Shoeb, Abubakar Abid, Adam Fisch, Adam~R Brown, Adam Santoro, Aditya Gupta, Adri{\`a} Garriga-Alonso, et~al. 2022.
\newblock Beyond the imitation game: Quantifying and extrapolating the capabilities of language models.
\newblock \emph{arXiv preprint arXiv:2206.04615}.

\bibitem[{Team(2023)}]{2023internlm}
InternLM Team. 2023.
\newblock Internlm: A multilingual language model with progressively enhanced capabilities.
\newblock \url{https://github.com/InternLM/InternLM}.

\bibitem[{Thirunavukarasu et~al.(2023)Thirunavukarasu, Hassan, Mahmood, Sanghera, Barzangi, El~Mukashfi, and Shah}]{thirunavukarasu2023trialling}
Arun~James Thirunavukarasu, Refaat Hassan, Shathar Mahmood, Rohan Sanghera, Kara Barzangi, Mohanned El~Mukashfi, and Sachin Shah. 2023.
\newblock Trialling a large language model (chatgpt) in general practice with the applied knowledge test: observational study demonstrating opportunities and limitations in primary care.
\newblock \emph{JMIR Medical Education}, 9(1):e46599.

\bibitem[{Touvron et~al.(2023{\natexlab{a}})Touvron, Lavril, Izacard, Martinet, Lachaux, Lacroix, Rozi{\`e}re, Goyal, Hambro, Azhar et~al.}]{touvron2023llama}
Hugo Touvron, Thibaut Lavril, Gautier Izacard, Xavier Martinet, Marie-Anne Lachaux, Timoth{\'e}e Lacroix, Baptiste Rozi{\`e}re, Naman Goyal, Eric Hambro, Faisal Azhar, et~al. 2023{\natexlab{a}}.
\newblock Llama: Open and efficient foundation language models.
\newblock \emph{arXiv preprint arXiv:2302.13971}.

\bibitem[{Touvron et~al.(2023{\natexlab{b}})Touvron, Martin, Stone, Albert, Almahairi, Babaei, Bashlykov, Batra, Bhargava, Bhosale et~al.}]{touvron2023llama2}
Hugo Touvron, Louis Martin, Kevin Stone, Peter Albert, Amjad Almahairi, Yasmine Babaei, Nikolay Bashlykov, Soumya Batra, Prajjwal Bhargava, Shruti Bhosale, et~al. 2023{\natexlab{b}}.
\newblock Llama 2: Open foundation and fine-tuned chat models.
\newblock \emph{arXiv preprint arXiv:2307.09288}.

\bibitem[{Wang et~al.(2018)Wang, Singh, Michael, Hill, Levy, and Bowman}]{wang2018glue}
Alex Wang, Amanpreet Singh, Julian Michael, Felix Hill, Omer Levy, and Samuel~R Bowman. 2018.
\newblock Glue: A multi-task benchmark and analysis platform for natural language understanding.
\newblock \emph{arXiv preprint arXiv:1804.07461}.

\bibitem[{Wang et~al.(2022)Wang, Wei, Schuurmans, Le, Chi, Narang, Chowdhery, and Zhou}]{wang2022self}
Xuezhi Wang, Jason Wei, Dale Schuurmans, Quoc Le, Ed~Chi, Sharan Narang, Aakanksha Chowdhery, and Denny Zhou. 2022.
\newblock Self-consistency improves chain of thought reasoning in language models.
\newblock \emph{arXiv preprint arXiv:2203.11171}.

\bibitem[{Wei et~al.(2022)Wei, Wang, Schuurmans, Bosma, Xia, Chi, Le, Zhou et~al.}]{wei2022chain}
Jason Wei, Xuezhi Wang, Dale Schuurmans, Maarten Bosma, Fei Xia, Ed~Chi, Quoc~V Le, Denny Zhou, et~al. 2022.
\newblock Chain-of-thought prompting elicits reasoning in large language models.
\newblock \emph{Advances in Neural Information Processing Systems}, 35:24824--24837.

\bibitem[{Wu et~al.(2023)Wu, Wang, Ye, Feng, Xu, Qiao, and Wu}]{wu2023openicl}
Zhenyu Wu, YaoXiang Wang, Jiacheng Ye, Jiangtao Feng, Jingjing Xu, Yu~Qiao, and Zhiyong Wu. 2023.
\newblock Openicl: An open-source framework for in-context learning.
\newblock \emph{arXiv preprint arXiv:2303.02913}.

\bibitem[{Xie et~al.(2021)Xie, Raghunathan, Liang, and Ma}]{xie2021explanation}
Sang~Michael Xie, Aditi Raghunathan, Percy Liang, and Tengyu Ma. 2021.
\newblock An explanation of in-context learning as implicit bayesian inference.
\newblock \emph{arXiv preprint arXiv:2111.02080}.

\bibitem[{Yang et~al.(2023)Yang, Hui, Yang, Li, Huang, and Li}]{yang2023iterative}
Jiaxi Yang, Binyuan Hui, Min Yang, Binhua Li, Fei Huang, and Yongbin Li. 2023.
\newblock Iterative forward tuning boosts in-context learning in language models.
\newblock \emph{arXiv preprint arXiv:2305.13016}.

\bibitem[{Yao et~al.(2023)Yao, Yu, Zhao, Shafran, Griffiths, Cao, and Narasimhan}]{yao2023tree}
Shunyu Yao, Dian Yu, Jeffrey Zhao, Izhak Shafran, Thomas~L Griffiths, Yuan Cao, and Karthik Narasimhan. 2023.
\newblock Tree of thoughts: Deliberate problem solving with large language models.
\newblock \emph{arXiv preprint arXiv:2305.10601}.

\bibitem[{Ye et~al.(2022)Ye, Iyer, Celikyilmaz, Stoyanov, Durrett, and Pasunuru}]{ye2022complementary}
Xi~Ye, Srinivasan Iyer, Asli Celikyilmaz, Ves Stoyanov, Greg Durrett, and Ramakanth Pasunuru. 2022.
\newblock Complementary explanations for effective in-context learning.
\newblock \emph{arXiv preprint arXiv:2211.13892}.

\bibitem[{Yuan et~al.(2023)Yuan, Yuan, Tan, Wang, and Huang}]{yuan2023well}
Zheng Yuan, Hongyi Yuan, Chuanqi Tan, Wei Wang, and Songfang Huang. 2023.
\newblock How well do large language models perform in arithmetic tasks?
\newblock \emph{arXiv preprint arXiv:2304.02015}.

\bibitem[{Zellers et~al.(2019)Zellers, Holtzman, Bisk, Farhadi, and Choi}]{zellers2019hellaswag}
Rowan Zellers, Ari Holtzman, Yonatan Bisk, Ali Farhadi, and Yejin Choi. 2019.
\newblock Hellaswag: Can a machine really finish your sentence?
\newblock \emph{arXiv preprint arXiv:1905.07830}.

\bibitem[{Zhong et~al.(2023)Zhong, Cui, Guo, Liang, Lu, Wang, Saied, Chen, and Duan}]{zhong2023agieval}
Wanjun Zhong, Ruixiang Cui, Yiduo Guo, Yaobo Liang, Shuai Lu, Yanlin Wang, Amin Saied, Weizhu Chen, and Nan Duan. 2023.
\newblock Agieval: A human-centric benchmark for evaluating foundation models.
\newblock \emph{arXiv preprint arXiv:2304.06364}.

\bibitem[{Zhou et~al.(2022)Zhou, Sch{\"a}rli, Hou, Wei, Scales, Wang, Schuurmans, Bousquet, Le, and Chi}]{zhou2022least}
Denny Zhou, Nathanael Sch{\"a}rli, Le~Hou, Jason Wei, Nathan Scales, Xuezhi Wang, Dale Schuurmans, Olivier Bousquet, Quoc Le, and Ed~Chi. 2022.
\newblock Least-to-most prompting enables complex reasoning in large language models.
\newblock \emph{arXiv preprint arXiv:2205.10625}.

\end{thebibliography}
\bibliographystyle{acl_natbib}

\appendix
\section{ICLEval Benchmark}
\label{sec:dataset samples}

\subsection{Data Source}
The data for our ICLEval Benchmark is sourced from various places, including Wikipedia, a common noun vocabulary, GSM8K, AQuA, BIGBench, and generation context from ChatGPT.

For paragraph-level and sentence-level data, we collected information from Wikipedia. Initially, we randomly selected 100 articles from Wikipedia to form the basis of natural language paragraphs. These paragraphs serve as the core content for the string completion task, which tests the models' ability to accurately copy information in unstructured context. It's important to note that we processed these paragraphs to ensure that the models hadn't encountered them during the pre-training stage. Additionally, we extracted all the sentences containing 5-30 words from these Wikipedia paragraphs to create sentence-level data. This data is used for tasks such as dictionary search, order adjustment, and de-duplication.

For word-level data, we gathered 1526 common nouns from a common noun vocabulary. Furthermore, we included all numbers, uppercase letters, and lowercase letters as character-level data. These datasets primarily focus on tasks related to solving order-related problems.

In addition, we sampled 50 examples from the GMS8K and AuQA datasets, respectively, to create the format cloning task. It's important to note that we are not concerned with the accuracy of the math questions themselves, but rather with the format of the predictions. Furthermore, we generated virtual data in the "person" and "company" domains using ChatGPT, which serves as the foundational content for tasks such as format check and format conversion.

For tasks related to statistics problems, we drew inspiration from the "navigate" and "long\_context\_integration" tasks in BIGBench\citep{srivastava2022beyond}. Additionally, we utilized the data from the "list\_numbers" task to create our numbers' rules task.

\subsection{Tasks}
We partition our tasks into two categories: testing copying abilities and testing learning abilities. The first two tasks focus on assessing copying abilities, while the remaining tasks evaluate learning abilities.

\begin{figure*}[!t]
    \centering
    \includegraphics[width=1\linewidth]{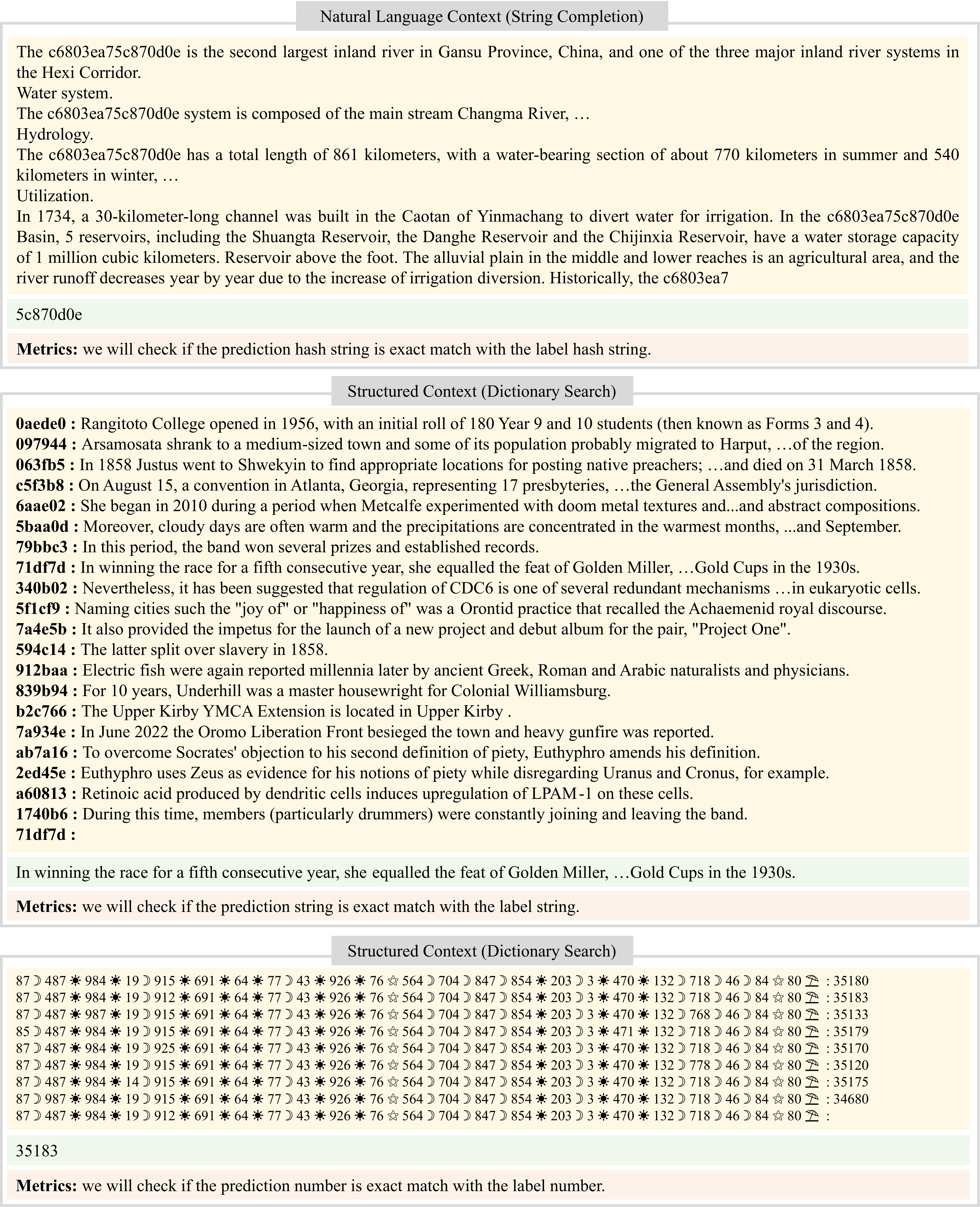}
    \caption{The tasks for copying ability evaluation. The prompt and label are in the yellow block and green block respectively. The metrics description is in the red block.}
    \label{fig:sample1}
\end{figure*}

\textbf{String Completion} task is specifically designed to test copying abilities in natural language context scenarios. In this task, models are required to predict the second half of a string given the first half. For each testing sample, we randomly select a lengthy paragraph from Wikipedia as the basis for a natural language context. To ensure that the model has not encountered the main entity mentioned in the paragraph before, we replace it with a 16-character hash string. Furthermore, we split the last hash string in the paragraph into two 8-character halves, and the target for the model is to predict the subsequent 8 characters. We have designed this task because it closely resembles the pre-training task and serves as a fundamental form of assessing copying ability. An example of this task can be seen in Figure~\ref{fig:sample1}.

\textbf{Dictionary Search} task involves predicting a value based on a given key from a set of key-value pairs. We explore two different forms of this task. The first form consists of short keys and long values, while the second form features short values but longer, more intricate keys. The samples in this task can be seen in Figure~\ref{fig:sample1}.

In the first form, we provide 20 key-value pairs with varying lengths as in-context examples. The key is a 6-character random hash string, while the value is a random sentence sampled from Wikipedia. 

The second form might be more difficult. We provide 10 key-value pairs that involve number calculations as in-context examples. The key is a long number calculation string with more than 20 numbers and operators. Moreover, to examine the impact of similar keys, we establish three levels for this task: "all-similar," "half-similar," and "non-similar." These levels determine the number of keys that share a similarity with the prediction target.

\begin{figure*}[!t]
    \centering
    \includegraphics[width=1\linewidth]{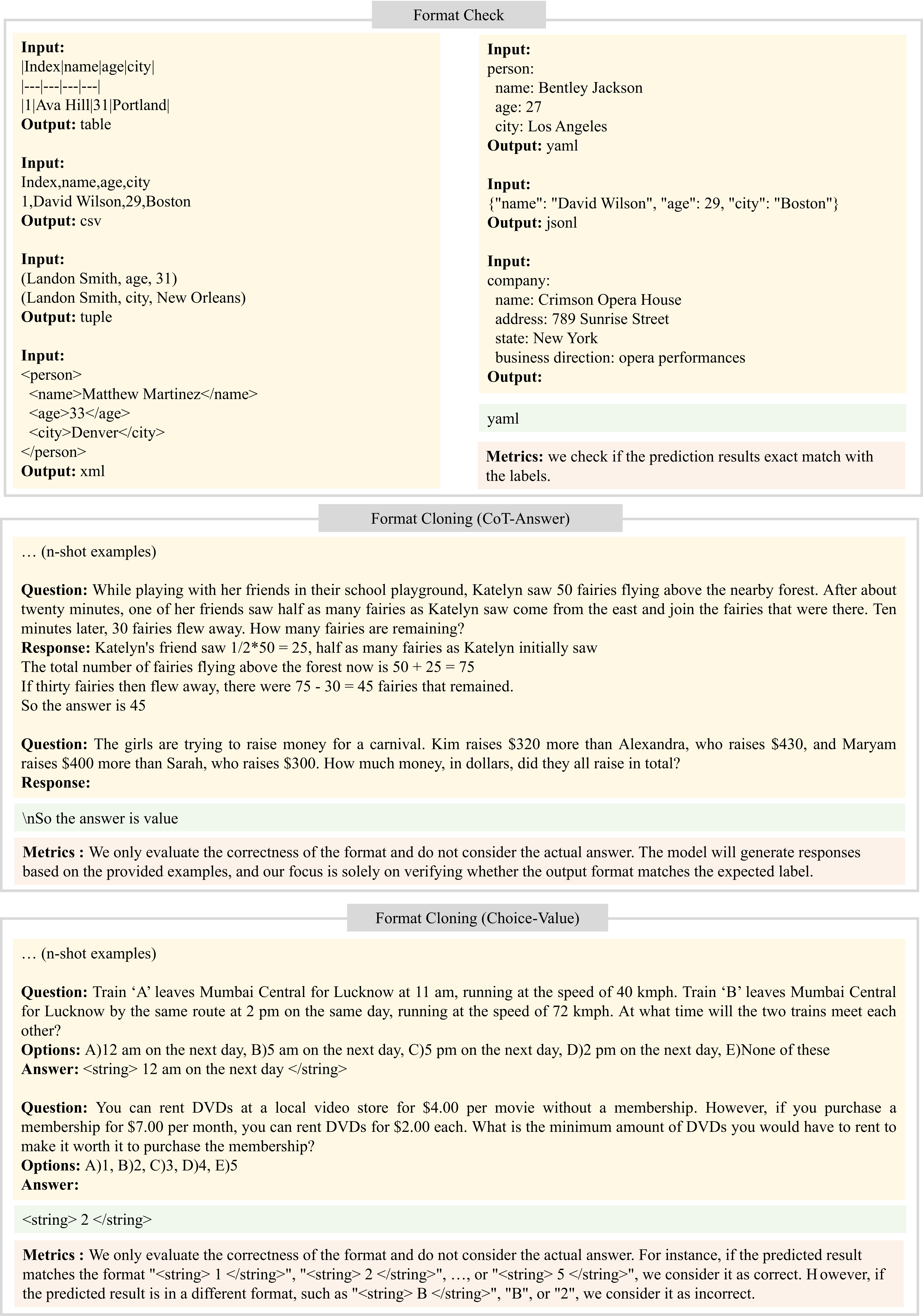}
    \caption{The samples of format check and format cloning tasks. The prompt and label are in the yellow block and green block respectively. The metrics description is in the red block.}
    \label{fig:sample2}
\end{figure*}

\textbf{Format Check} is a classification task to distinguish which class the current format is. We chose six common formats: JSONL, CSV, Triple-Tuple, YAML, Markdown-Table, and XML. Then we will randomly generate 6-shot examples using the six formats respectively as the in-context for each testing sample. Each sample is a random format and belongs to the six formats. The samples in this task can be seen in Figure~\ref{fig:sample2}.

\textbf{Format Cloning} is a generation task to follow the customized formats. We have designed five customized formats for the CoT output of GSM8K \citep{cobbe2021training} and the multi-choice output of AQuA \citep{ling2017program}. And we randomly sample 5-shot examples from this dataset. In these tasks, we check the correctness of the output format without focusing on the correctness of the results themselves. The samples in this task can be seen in Figure~\ref{fig:sample2}.

\begin{figure*}[!t]
    \centering
    \includegraphics[width=1\linewidth]{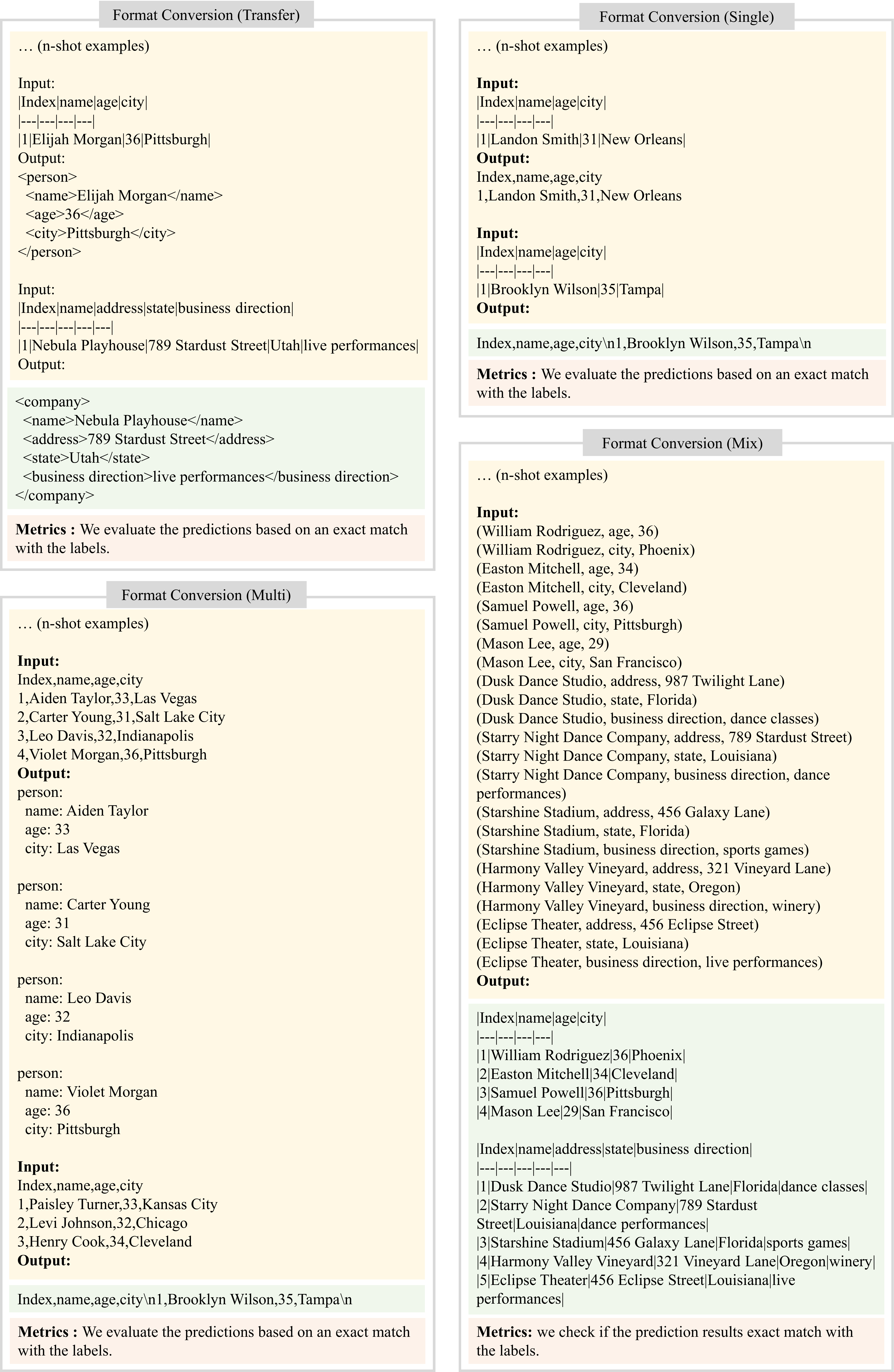}
    \caption{The samples of format conversion tasks with four different forms: "single", "multi", "transfer" and "mix".}
    \label{fig:sample3}
\end{figure*}

\textbf{Format Conversion} is a generation task to convert a source format to a target format and keep the content consistent. We also use the six formats mentioned in the "format-check" task. To explore more complex scenarios, we set four forms in this task: "single", “multi", "transfer" and "mix". We have two domains ("people" and "company") of data as content. "single" form means only converting one item, and the domain is consistent between in-context examples and testing samples. "multi" form will have 1-5 items based on the "single" form. While the "transfer" form also converts one item the in-context examples and testing samples are from different domains. The "mix" form might most difficult which means there will be multiple items and both two domains' data as content. The samples in this task can be seen in Figure~\ref{fig:sample3}.

\begin{figure*}[!t]
    \centering
    \includegraphics[width=1\linewidth]{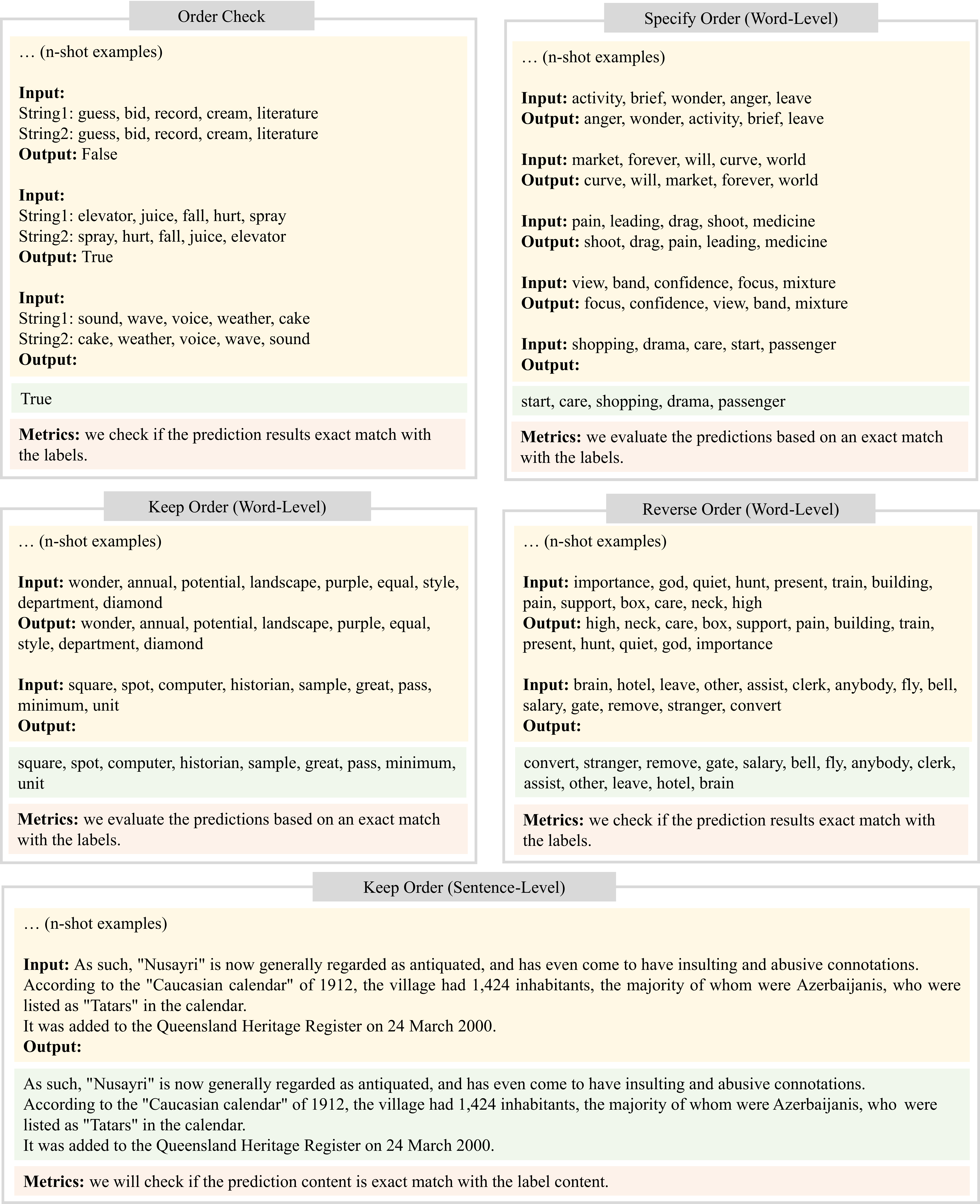}
    \caption{The samples of order check and order adjustment tasks.  The prompt and label are in the yellow block and green block respectively. The metrics description is in the red block.}
    \label{fig:sample4}
\end{figure*}

\textbf{Order Check} is a classification task to judge if the order of elements in one string is reversed. We set 8-shot examples and each example has two strings. If the two strings have the same order, the label is False, while if the two strings have reversed order, the label is True. We will randomly generate different 8-shot examples for every testing sample, and the proportion of positive and negative examples is also random. We have word-level and character-level settings for different element granularity. The sample in this task can be seen in Figure~\ref{fig:sample4}.

\textbf{Order Adjustment} is a generation task to generate sequence, reverse, or specify form string given origin string. For the sequence setting, the model is only required to copy the input text without any changes. For the reverse sequence setting, the model needs to output the reversed version of the original text. For the specific sequence setting, the model must learn to identify and execute a specified index operation on the input text. We randomly generate 5-shot examples for each testing sample, and hope models can learn and apply the order adjustment rules to the testing sample. To enhance diversity, we experiment with character-level, word-level, and sentence-level input-output pairs for different element granularity. The sample in this task can be seen in Figure~\ref{fig:sample4}.

\begin{figure*}[!t]
    \centering
    \includegraphics[width=1\linewidth]{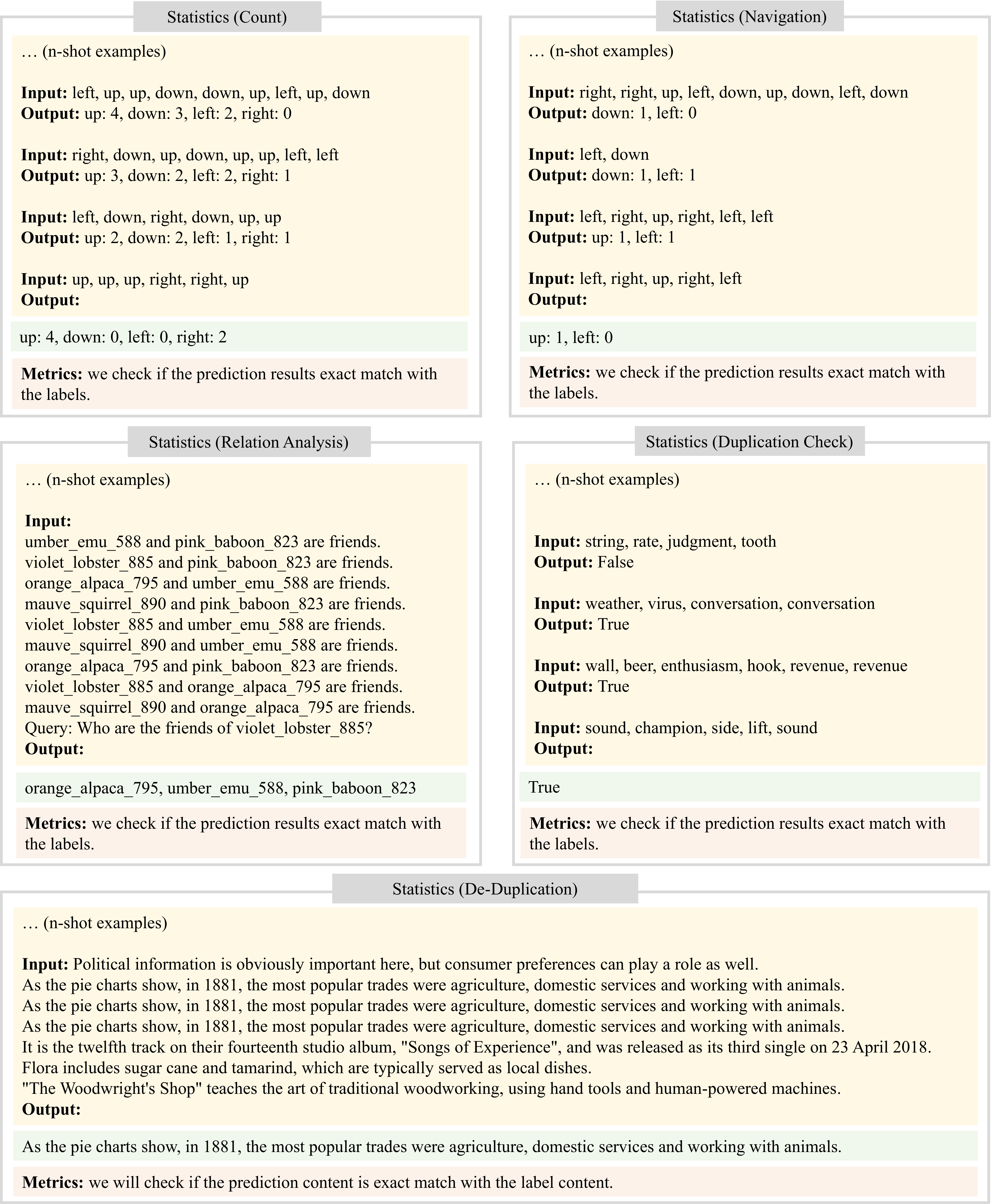}
    \caption{The samples of tasks about statistics problem. The prompt and label are in the yellow block and green block respectively. The metrics description is in the red block.}
    \label{fig:sample5}
\end{figure*}

\textbf{Count \& Navigation} is a generation task to return a dictionary depending on the in-context information. The model is given a list of dictionary nouns (up, down, right, and left) as input. The count task requires the model to count the number of occurrences of all the nouns, while the navigation task involves considering the final point relative to the initial point. We set "easy" mode and "hard" mode for this task. The "easy" mode only has two dictionary nouns while the "hard" mode has four. The sample in this task can be seen in Figure~\ref{fig:sample5}.

\textbf{Relation Analysis} task provides the model with a relation graph as input and requires it to output all the nodes connected to a chosen node. Each of these tasks includes 5-shot examples as in-context information. We have adjusted the nodes or sides in the relation graph for this task, to introduce variations for different models. The sample in this task can be seen in Figure~\ref{fig:sample5}.

\textbf{Duplication Check} a classification task to judge if there are repeated elements in the string. If there are repeated elements in the string, the label is True, else the label is False. We use the same setting as the "order-check" task, which has 8-shot examples for every testing sample and has word-level and character-level element granularity. The sample in this task can be seen in Figure~\ref{fig:sample5}.

\textbf{De-Duplication} is a generation task to find or remove repeat elements from a string. Finding repeat elements means only outputting the repeat element in one string while removing repeat elements means outputting a whole string without any repeat elements. We use sentence-level settings for finding repeat elements form and use character-level and word-level settings for removing repeat elements form. As previous setting, we randomly generate 5-shot examples for each testing sample. The sample in this task can be seen in Figure~\ref{fig:sample5}.

\textbf{Numbers' List Mapping} is a task that consists of multiple groups of number pairs. Each pair comprises an input list and an output list, with an internal rule governing the transformation from input to output. With multi-group examples as the in-context information, the model needs to learn the underlying rule and predict the output list for a new input list. We have collected task data for this evaluation from the "list\_functions" task in BIGBench. In this task, there are 250 different rules, and some rules are even very hard for humans to find. The samples in this task can be seen in Figure~\ref{fig:bad-case-5}.

\begin{table}[!htbp]
    \centering
    \small
    \setlength\tabcolsep{1mm}{
    \begin{tabular}{lrc}
        \toprule
        \textbf{Task Name} & \textbf{Split String} & \textbf{Max Len.} \\
        \midrule
        \textbf{Copying} & & \\
        entity complete & all punctuation marks & 10\\
        dictionary search & & \\
        - short key & \textbackslash n & 75\\
        - long and similar key & \textbackslash n & 12\\
        \midrule
        \textbf{Learning} & & \\
        format check & \textbackslash n & 5 \\
        format cloning & \textbackslash nQuestion: & 196\\
        format conversion & \textbackslash nInput: & 256\\
        
        order check & \textbackslash n & 5\\
        order adjustment & &  \\
        - character-level & \textbackslash n & 50 \\
        - word-level & \textbackslash n & 50 \\
        - sentence-level & \textbackslash nInput: & 256  \\

        count \& navigation & \textbackslash n & 30\\
        relation analysis & \textbackslash nInput: & 60\\
        
        duplication check & \textbackslash n & 5\\
        de-duplication & &  \\
        - character-level & \textbackslash n & 30 \\
        - word-level & \textbackslash n & 30 \\
        - sentence-level & \textbackslash n & 60  \\
        
        numbers' rules & \textbackslash n & 50\\
        \bottomrule
    \end{tabular}}
    \caption{The max generation length and split string for different tasks. We set the max generation length for testing effectiveness.}
    \label{tab:test-settings}
\end{table} 

\subsection{Metrics}
\label{sec:metrics}
Considering the effectiveness of \ourdata, we set different max generation length limitations for various tasks. Meanwhile, we set different split strings for processing models' responses to final predictions. Both the settings are shown in Table~\ref{tab:test-settings}. Setting the two parameters is because the base-version model will not stop when giving in-context examples. The max generation length will make tested models stop their generation in the suit position. And we will regard the content in front of the split strings as the right predictions. Then we will strip the blank space at both ends of the right predictions.

\section{Evaluated Models}
\label{sec:models}
We show the details of our evaluated models in this section, and the models are listed in Table~\ref{tab:models}.

\begin{table}[!t]
    \centering
    \small
    \setlength\tabcolsep{1.5mm}{
    \begin{tabular}{lcrr}
        \toprule
        \textbf{Model} & \textbf{Version} & \textbf{Parameters} & \textbf{Training Data}\\
        \midrule
        GPT-Neo & Base & 1.3B & 0.38T Tokens\\
        Phi-1.5 & Base & 1.3B & -- \\
        TinyLlama & Base & 1.1B & 3T Tokens \\
        GPT-J  & Base & 6B & 0.4T Tokens \\
        \midrule
        \multirow{4}{*}{LLaMA1} & \multirow{4}{*}{Base} 
        & 7B & 1T Tokens \\
        & & 13B & 1T Tokens\\
        & & 34B & 1.4T Tokens\\
        & & 65B & 1.4T Tokens\\
        \multirow{3}{*}{LLaMA2} & \multirow{2}{*}{Base/Chat}
        & 7B & 2T Tokens \\
        & & 13B & 2T Tokens\\
        & {Base} & 70B & 2T Tokens\\
        \multirow{2}{*}{LLaMA3} & \multirow{2}{*}{Base}
        & 8B & >15T Tokens \\
        & & 70B & -- \\
        \midrule
        Mistral & {Base/Chat} & 7B & --\\
        Mistral-MoE & {Chat} & 8\(\times\)7B & --\\
        ChatGLM3 & Base/Chat & 6B & -- \\
        \multirow{2}{*}{YI} & \multirow{2}{*}{Base/Chat} & 6B & 3T Tokens\\
        & & 34B & 3T Tokens\\
        
        \multirow{2}{*}{Baichuan2} & \multirow{2}{*}{Base/Chat}
        & 7B & 2.6T Tokens \\
        & & 13B & 2.6T Tokens \\
        
        \multirow{2}{*}{Qwen} &
        \multirow{2}{*}{Base/Chat}
        & 7B & >2.4T Tokens\\
        & & 14B & >3T Tokens\\
        \midrule
        
        InternLM & {Base/Chat} & 7B & >1T Tokens\\

        Skywork & {Base} & 13B & 3.2T Tokens\\
        
        Amber & {Base} & 7B & 1.2T Tokens\\
        \bottomrule
    \end{tabular}}
    \caption{Models evaluated in our work}
    \label{tab:models}
\end{table}

\textbf{GPT-Neo-1.3B} is a transformer model with 1.3 billion parameters, trained on the Pile, a curated dataset created by EleutherAI specifically for training this model.

\textbf{Phi-1.5} is a transformer model with 1.3 billion parameters, trained by Microsoft Research. It is specialized in basic Python coding and is augmented with various NLP synthetic texts.

\textbf{TinyLlama} project \citep{tinyllama} aims to pretrain a 1.1 billion parameter Llama model on 3 trillion tokens. The model's intermediate checkpoints, corresponding to about 10 billion tokens, are open-sourced every 5,000 steps. This project is released by the StatNLP Research Group of Singapore University of Technology and Design.

\textbf{GPT-J} is a transformer model trained with 6 billion parameters. It is trained by EleutherAI.

\textbf{ChatGLM3-6B} is the latest open-source model in the ChatGLM series, utilizing the General Language Model (GLM) architecture with 6 billion parameters. It was released by the Knowledge Engineering Group (KEG) \& Data Mining at Tsinghua University.

\textbf{YI-6B and YI-34B} belong to the Yi series models, trained from scratch by 01.AI. These models are based on a 3 trillion multilingual corpus and have 6 billion and 34 billion parameters, respectively.

\textbf{LLaMA1-7B and LLaMA1-13B} belong to the Llama series models released by Meta AI. They are trained on 1 trillion tokens.

\textbf{LLaMA1-30B and LLaMA1-65B} belong to the Llama series models released by Meta AI. They are trained on 1.4 trillion tokens.

\textbf{LLaMA2-7B, LLaMA2-13B, and LLaMA2-70B} belong to the Llama 2 series models, which are a collection of pre-trained and fine-tuned generative text models ranging in scale from 7 billion to 70 billion parameters. These models are auto-regressive language models trained on 2 trillion tokens and were released by Meta AI.

\textbf{Mistral-7B-v0.1} is a pre-trained generative text model with 7 billion parameters. It was released by Mistral AI.

\textbf{Skywork-13B} was trained on a high-quality cleaned dataset consisting of 3.2 trillion multilingual data, mainly in Chinese and English, including code. It was released by Skywork AI.

\textbf{Baichuan2-7B and Baichuan2-13B} belong to the Baichuan2 series models, which are open-source and commercially usable large-scale language models developed by Baichuan Intelligence. They are trained on a high-quality corpus with 2.6 trillion tokens. Baichuan Intelligence also open-sources intermediate checkpoints of Baichuan2-7B every 220 billion tokens.

\textbf{Qwen-7B and Qwen-14B} are versions of the large language model series called Qwen (Tongyi Qianwen) proposed by Alibaba Cloud. These Transformer-based large language models have 7 billion, 14 billion, and 72 billion parameters, respectively. They are trained on a large volume of data, including web texts, books, code, etc.

\textbf{InternLM-7B} \citep{2023internlm} is an open-sourced base model with 7 billion parameters, tailored for practical scenarios.

\textbf{Amber-7B} is a 7 billion parameter language model with the same architecture as LLaMA-7B, trained on Arxiv, Book, C4, Refined-Web, StarCoder, StackExchange, and Wikipedia. It was released by LLM360 \citep{liu2023llm360}, which is an initiative for comprehensive and fully open-source language models. And it provides open-source 360 intermediate checkpoints of Amber-7B.

\section{Experiments Results}

\begin{figure}[!t]
    \centering
    \includegraphics[width=1\linewidth]{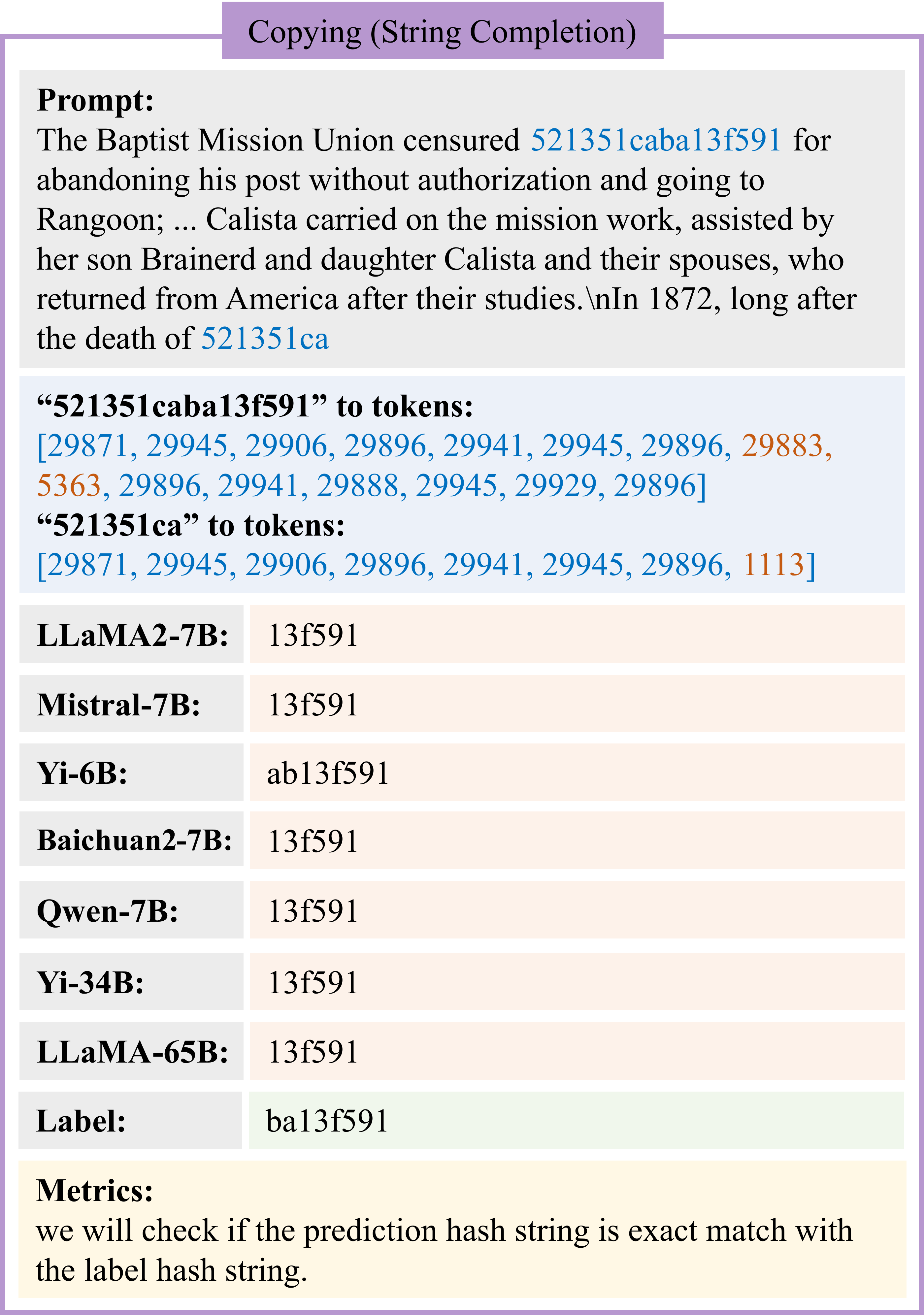}
    \caption{A bad case of entity completion task and order adjustment task. These two bad cases are selected from the predictions of LLaMA2-7B. }
    \label{fig:bad-case-1}
\end{figure}

\subsection{Bad Cases Analysis}
\label{sec:bad case}
We list some bad cases in this sub-section. 

Many models can arrive at about $0.89$ scores in the entity completion task, but it's hard for them to get higher scores. We analyzed the bad case and found that there are some entities (hash string) that have token-fracture phenomenons when splitting them. As shown in Figure~\ref{fig:bad-case-1}, the tokens of string “521351caba13f591” are not equal to the tokens of “521351ca" add the tokens of "ba13f591”. This makes it difficult for models to finish this task.

\begin{figure*}[!t]
    \centering
    \includegraphics[width=1\linewidth]{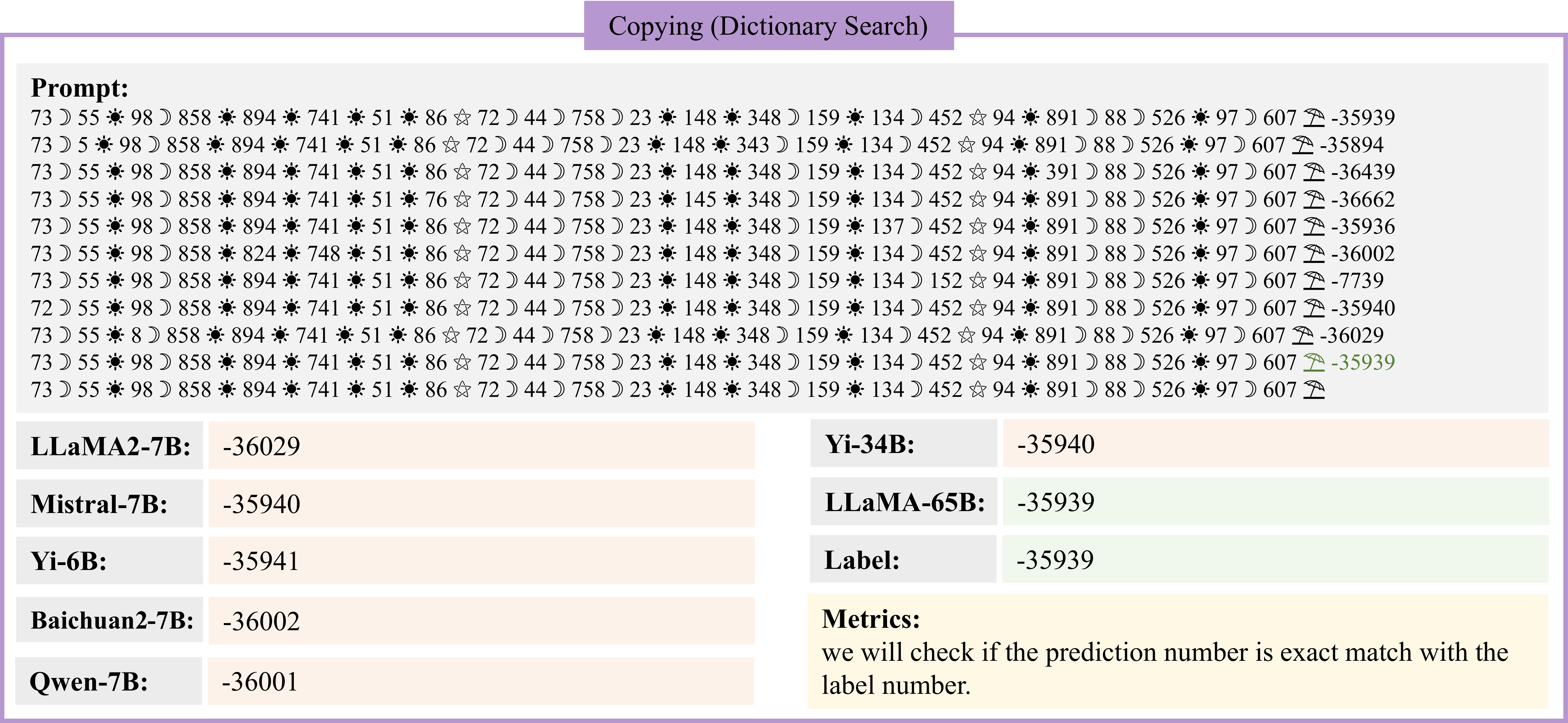}
    \caption{A bad case of dictionary search task (with all-similar examples).}
    \label{fig:bad-case-2}
\end{figure*}

We give a bad case of dictionary search task in Figure~\ref{fig:bad-case-2}. Similar examples confuse the models, and only the LLaMA-65B model finds the right value.

\begin{figure*}[!t]
    \centering
    \includegraphics[width=1\linewidth]{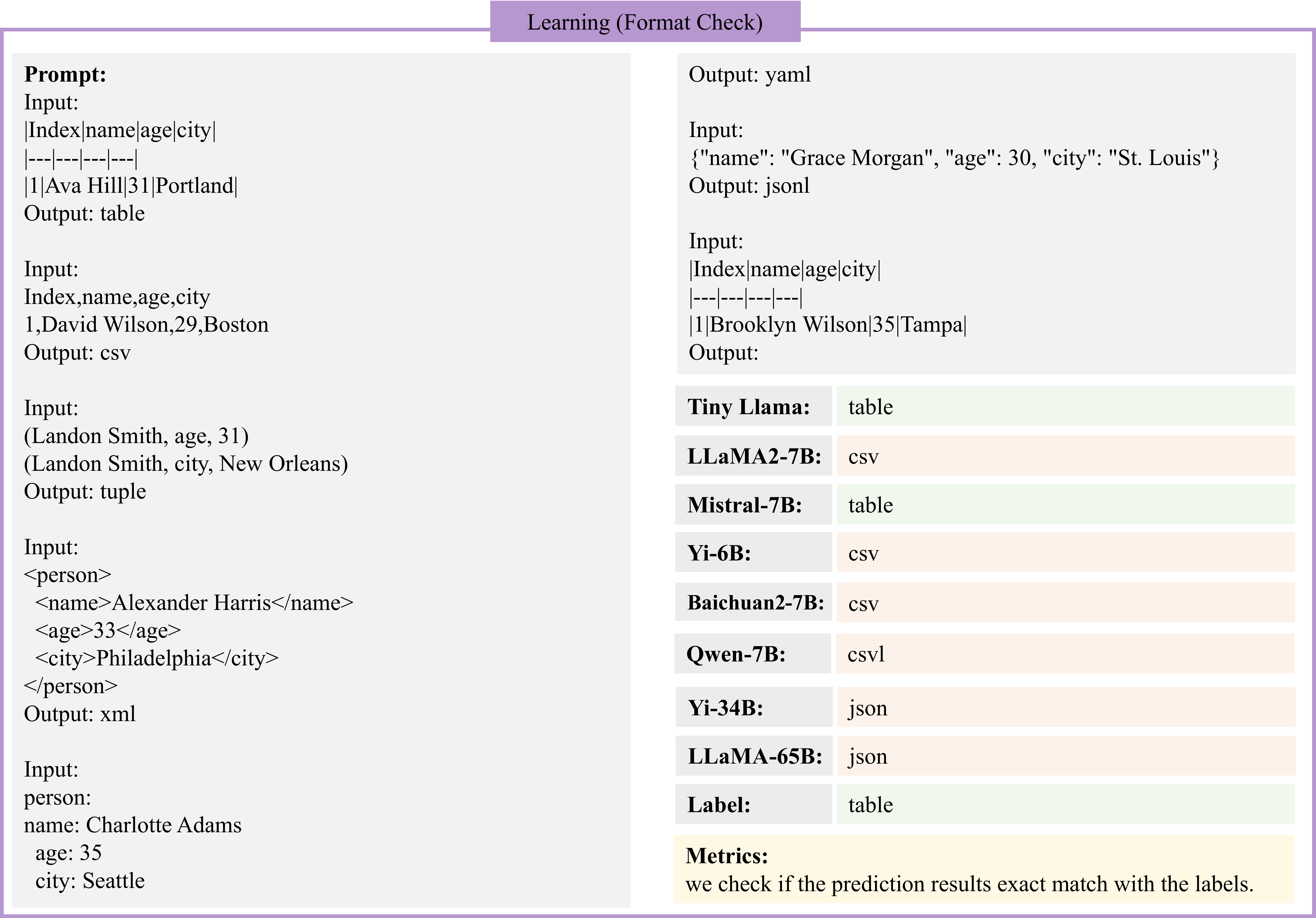}
    \caption{A bad case of format check task.}
    \label{fig:bad-case-3}
\end{figure*}

We give a bad case of format check task in Figure~\ref{fig:bad-case-3}. Lots of models predict the results depending on the models' inherent preferences rather than the in-context examples. The rules in this task are easy to learn so that small-sized models (e.g. TinyLlama) can predict the right labels. However, the larger models (e.g. LLaMA-65B) might predict wrong labels due to their inherent preferences.

\begin{figure*}[!t]
    \centering
    \includegraphics[width=1\linewidth]{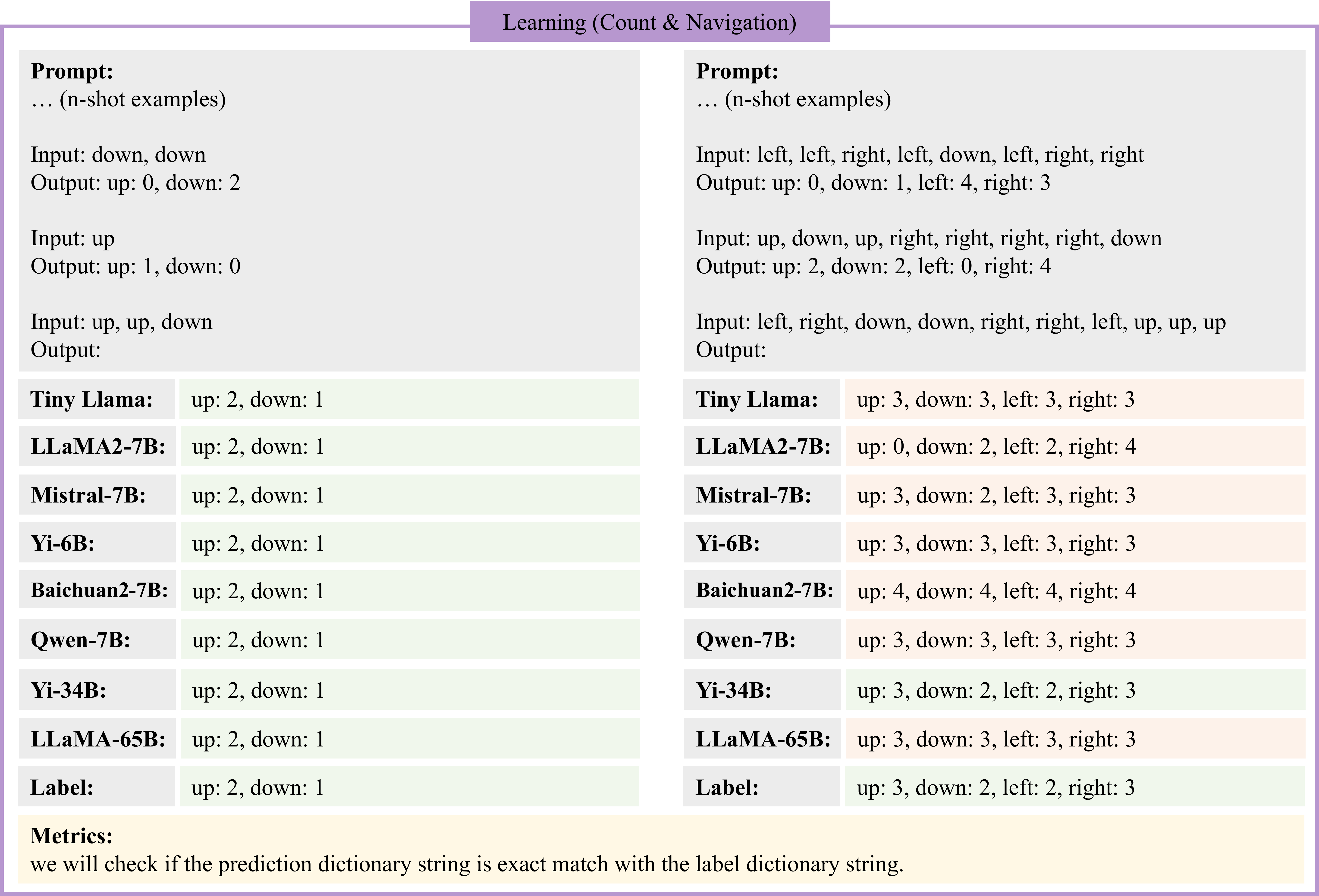}
    \caption{The bad cases of Count \& Navigation task.}
    \label{fig:bad-case-4}
\end{figure*}

We give two bad cases of count \& navigation task in Figure~\ref{fig:bad-case-4}. Models have good performance when there are several elements to statistics. The performance has a huge drop when the number of elements becomes greater.

\begin{figure*}[!t]
    \centering
    \includegraphics[width=1\linewidth]{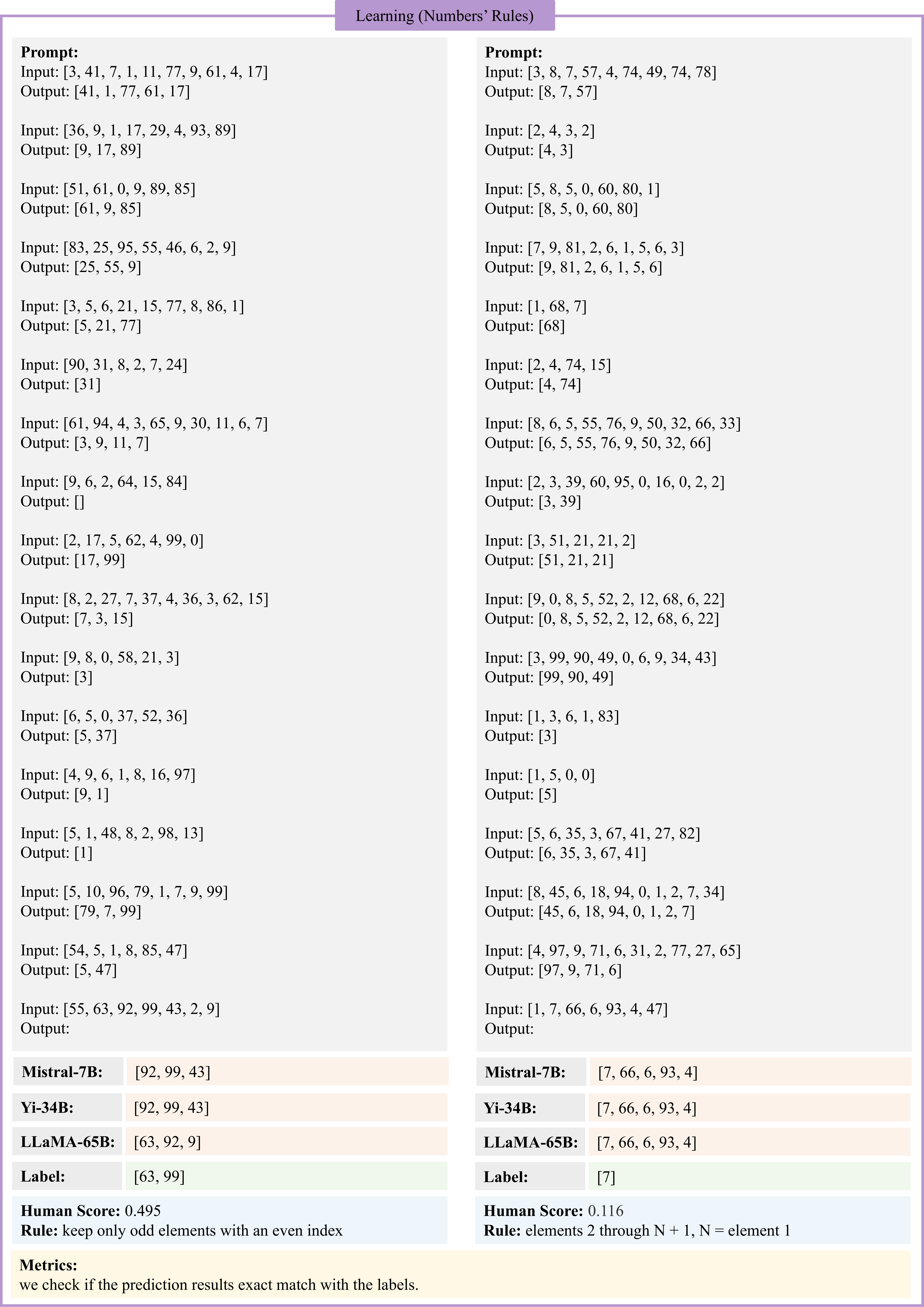}
    \caption{The bad cases of numbers' rules task.}
    \label{fig:bad-case-5}
\end{figure*}

We give two bad cases of numbers' rules task in Figure~\ref{fig:bad-case-5}. Some rules are even very difficult for humans. We chose two hard samples and showed the mean human score in this figure.

\subsection{Detailed Results}
\label{sec:detailed-results}
We list the detailed results in Table~\ref{tab:detailed-results}. Each column is the results of all the models we tested for one task. We list the mean results of all the tasks in the last column of this table. We also put the maximum and minimum results in the last two rows of the table.

\subsection{Results in Pretraining Stage}
\label{sec:pretraining-tokens}

\begin{figure*}[!t]
    \centering
    \includegraphics[width=1\linewidth]{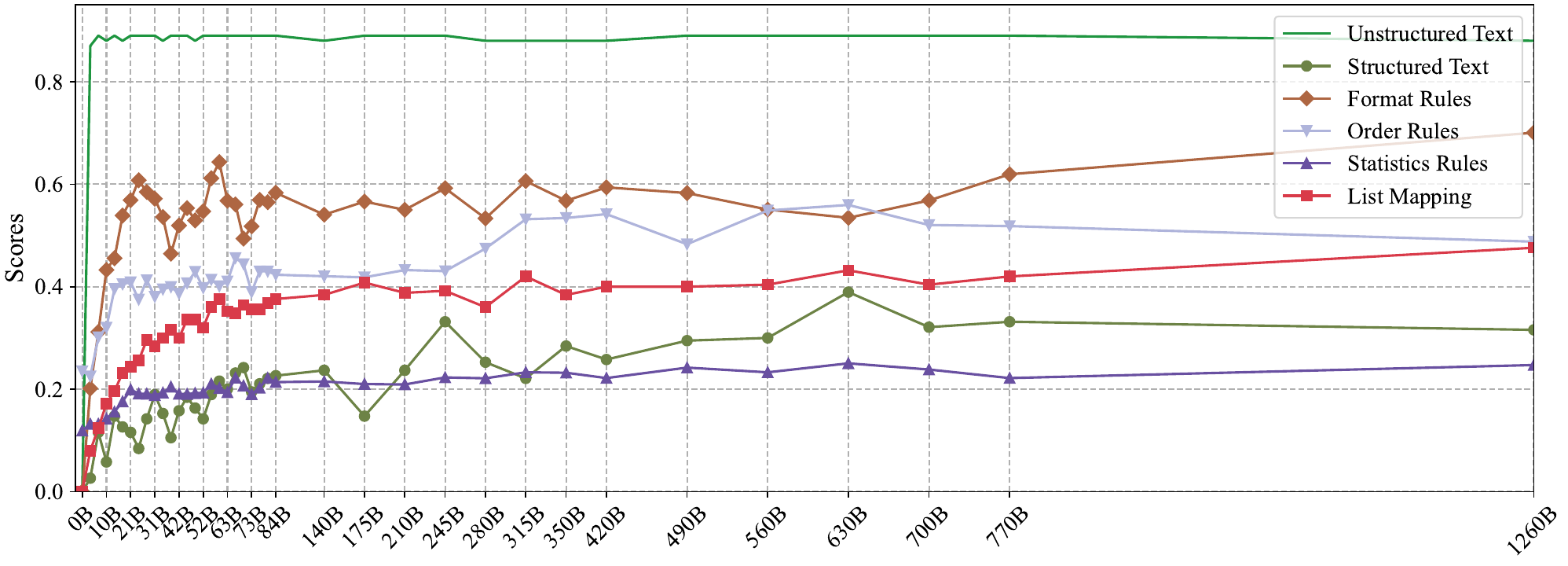}
    \caption{The pretrain stage of Amber-7B.}
    \label{fig:amber-pretraining-res}
\end{figure*}

In this section, we tested the Amber-7B models with the multiple checkpoints, trained on 1.2T tokens, and saved about every 3.5B tokens. The ICL abilities scores are shown in Figure~\ref{fig:amber-pretraining-res}. And we can get similar conclusions as the results of Baichuan2-7B and TinyLlama-1.1B.

\subsection{Influences of Evaluation Settings.}
\label{sec: evaluation settings}

Different evaluation settings might have can affect the results of ICL abilities, such as the order, the prompt, and the number of examples.

(1) We dynamically sample every example for each individual sample within our tasks, as opposed to using a fixed set of examples for all samples. Therefore, we believe that settings effectively address the issue of order.

(2) We only use the “input-output” as the prefix for all the examples in our main experiment. We try some experiments to set different prompts, such as “Question-Answer”, “123-456”, “Q-A”, and “answer-question” to replace our original prompt “Input-Output”, as shown in Table~\ref{tab:adjust-prompt}. The results are shown in the table below. We can see that 1) the prompt with influence the models’ results when the number of examples is not enough, but has little influence when the number of examples rises. 2) the prompt settings nearly do not influence the relative results between different models. So, there is reason to believe that the prompt setting will not influence the conclusion of our main experiment. In considering the robust evaluation, we will add different prompts in our testing data in the future.

(3) More examples will bring better results. We select the suitable n-shot settings for greater discrimination in our main experiment. We also adjusted the n-shot settings for several tasks (format checking and count \& navigation), and the results are shown in Table~\ref{tab:adjust-n-shot-format} and Table~\ref{tab:adjust-n-shot-count}.

\begin{table}[!t]
    \centering
    \small
    \setlength\tabcolsep{1.8mm}{
    \begin{tabular}{lccccc}
        \toprule
        \multirow{2}{*}{\textbf{N-shot}} & \multicolumn{2}{c}{\textbf{LLaMA2-7B}} & \multicolumn{2}{c}{\textbf{LLaMA2-13B}} & \multirow{2}{*}{\textbf{Mean}}\\
        & normal & transfer & normal & transfer & \\
        \midrule
        6-shot	&0.017	&0.300	&0.000	&0.233	&0.138 \\
        12-shot	&0.983	&1.000	&0.983	&0.967	&0.983 \\
        24-shot	&1.000	&0.983	&1.000	&1.000	&0.996 \\
        \bottomrule
    \end{tabular}}
    \caption{Different n-shot settings in Format Checking Task. "normal" and "transfer" are two different subsets in this task.}
    \label{tab:adjust-n-shot-format}
\end{table}

\begin{table*}[!t]
    \centering
    \small
    \setlength\tabcolsep{0.8mm}{
    \begin{tabular}{lccccccccccccc}
        \toprule
        \multirow{2}{*}{\textbf{Models}} &String  &Dict.  &Order  &Order &Dupl. &De- &Rel. &Count &Format  &Format  &Format &List & \multirow{2}{*}{\textbf{Mean}}\\
        & Comp. & Search & Check & Adjust. &  Check & Dupl. & Anal. & \& Nav. & Check & Cloning & Conv. & Map.& \\
        \midrule
        \textbf{Random}& 0.000 & 0.000 & 0.500 & 0.000 & 0.500 & 0.000 & 0.000 & 0.000 & 0.167 & 0.000 & 0.000 & 0.000 & 0.072\\
        \midrule
        TinyLlama-1.1B& 0.458 & 0.890 & 0.540 & 0.379 & 0.557 & 0.223 & 0.010 & 0.125 & 0.375 & 0.827 & 0.592 & 0.356 & 0.498\\
        Phi-1.5& 0.253 & 0.540 & 0.550 & 0.379 & 0.573 & 0.287 & 0.030 & 0.100 & 0.175 & 0.720 & 0.550 & 0.388 & 0.396\\
        GPT-Neo-1.3B& 0.479 & 0.540 & 0.610 & 0.387 & 0.533 & 0.180 & 0.010 & 0.092 & 0.175 & 0.777 & 0.492 & 0.376 & 0.430\\
        \midrule
        GPT-J& 0.463 & 0.530 & 0.730 & 0.438 & 0.637 & 0.233 & 0.030 & 0.133 & 0.617 & 0.833 & 0.675 & 0.492 & 0.506\\
        ChatGLM3-6B& 0.700 & 0.900 & 0.750 & 0.333 & 0.517 & 0.273 & 0.170 & 0.317 & 0.483 & 0.840 & 0.475 & 0.432 & 0.582\\
        ChatGLM3-6B-Chat& 0.342 & 0.890 & 0.610 & 0.363 & 0.550 & 0.327 & 0.100 & 0.225 & 0.675 & 0.887 & 0.725 & 0.372 & 0.526\\
        Yi-6B& 0.379 & 0.870 & 0.740 & 0.488 & 0.500 & 0.340 & 0.070 & 0.217 & 0.100 & 0.943 & 0.792 & 0.492 & 0.541\\
        Yi-6B-Chat& 0.484 & 0.870 & 0.640 & 0.525 & 0.623 & 0.447 & 0.140 & 0.258 & 0.033 & 0.863 & 0.808 & 0.452 & 0.554\\
        Yi-34B& 0.600 & 0.860 & 0.980 & 0.696 & 0.730 & 0.627 & 0.470 & 0.525 & 0.117 & 0.960 & 0.808 & 0.536 & 0.675\\
        Skywork-13B& 0.458 & 0.880 & 0.930 & 0.438 & 0.657 & 0.357 & 0.130 & 0.350 & 0.492 & 0.953 & 0.808 & 0.472 & 0.603\\
        InternLM-7B& 0.421 & 0.550 & 0.630 & 0.404 & 0.577 & 0.277 & 0.100 & 0.158 & 0.317 & 0.877 & 0.767 & 0.456 & 0.479\\
        InternLM-7B-Chat& 0.732 & 0.530 & 0.640 & 0.308 & 0.620 & 0.000 & 0.000 & 0.258 & 0.575 & 0.427 & 0.425 & 0.380 & 0.468\\
        \midrule
        Qwen-7B& 0.453 & 0.890 & 0.750 & 0.487 & 0.687 & 0.523 & 0.150 & 0.333 & 0.100 & 0.927 & 0.700 & 0.484 & 0.574\\
        Qwen-7B-Chat& 0.721 & 0.890 & 0.630 & 0.354 & 0.643 & 0.337 & 0.110 & 0.300 & 0.083 & 0.813 & 0.342 & 0.492 & 0.559\\
        Qwen-14B& 0.484 & 0.870 & 0.870 & 0.717 & 0.707 & 0.637 & 0.350 & 0.467 & 0.308 & 0.923 & 0.808 & 0.552 & 0.653\\
        Qwen-14B-Chat& 0.584 & 0.900 & 0.770 & 0.683 & 0.720 & 0.623 & 0.350 & 0.467 & 0.167 & 0.546 & 0.758 & 0.552 & 0.632\\
        \midrule
        Baichuan2-7B& 0.442 & 0.870 & 0.610 & 0.454 & 0.603 & 0.337 & 0.110 & 0.308 & 0.158 & 0.933 & 0.800 & 0.492 & 0.551\\
        Baichuan2-7B-Chat& 0.516 & 0.880 & 0.670 & 0.429 & 0.557 & 0.403 & 0.140 & 0.233 & 0.058 & 0.773 & 0.642 & 0.456 & 0.538\\
        Baichuan2-13B& 0.232 & 0.800 & 0.870 & 0.450 & 0.617 & 0.467 & 0.210 & 0.325 & 0.633 & 0.893 & 0.725 & 0.392 & 0.540\\
        Baichuan2-13B-Chat& 0.363 & 0.750 & 0.830 & 0.396 & 0.630 & 0.510 & 0.270 & 0.392 & 0.575 & 0.887 & 0.683 & 0.356 & 0.541\\
        \midrule
        Mistral-7B& 0.537 & 0.880 & 0.980 & 0.575 & 0.623 & 0.490 & 0.300 & 0.442 & 0.217 & 0.947 & 0.842 & 0.560 & 0.648\\
        Mistral-7B-Chat& 0.763 & 0.870 & 0.720 & 0.471 & 0.603 & 0.493 & 0.190 & 0.358 & 0.450 & 0.917 & 0.833 & 0.468 & 0.640\\
        Mixtral-8x7B-Chat& 0.884 & 0.880 & 0.990 & 0.737 & 0.793 & 0.677 & 0.530 & 0.517 & 0.358 & 0.953 & 0.833 & 0.592 & 0.761\\
        \midrule
        LLaMA-7B& 0.332 & 0.870 & 0.670 & 0.412 & 0.587 & 0.327 & 0.030 & 0.150 & 0.358 & 0.900 & 0.775 & 0.416 & 0.518\\
        LLaMA-13B& 0.426 & 0.890 & 0.650 & 0.442 & 0.577 & 0.340 & 0.080 & 0.275 & 0.192 & 0.907 & 0.800 & 0.472 & 0.547\\
        LLaMA-30B& 0.332 & 0.890 & 0.910 & 0.629 & 0.637 & 0.447 & 0.350 & 0.425 & 0.475 & 0.940 & 0.783 & 0.528 & 0.619\\
        LLaMA-65B& 0.495 & 0.890 & 0.970 & 0.729 & 0.613 & 0.563 & 0.330 & 0.392 & 0.000 & 0.947 & 0.833 & 0.576 & 0.646\\
        LLaMA2-7B& 0.453 & 0.890 & 0.830 & 0.429 & 0.630 & 0.293 & 0.120 & 0.233 & 0.158 & 0.917 & 0.767 & 0.452 & 0.560\\
        LLaMA2-7B-Chat& 0.758 & 0.890 & 0.730 & 0.454 & 0.623 & 0.403 & 0.220 & 0.317 & 0.258 & 0.920 & 0.650 & 0.412 & 0.609\\
        LLaMA2-13B& 0.474 & 0.890 & 0.830 & 0.608 & 0.620 & 0.423 & 0.090 & 0.308 & 0.117 & 0.947 & 0.750 & 0.520 & 0.595\\
        LLaMA2-13B-Chat& 0.747 & 0.900 & 0.750 & 0.604 & 0.620 & 0.467 & 0.260 & 0.292 & 0.133 & 0.900 & 0.658 & 0.488 & 0.631\\
        LLaMA2-70B& 0.432 & 0.890 & 0.980 & 0.758 & 0.747 & 0.687 & 0.470 & 0.450 & 0.483 & 0.933 & 0.842 & 0.580 & 0.685\\
        LLaMA3-8B& 0.874 & 0.570 & 0.980 & 0.892 & 0.670 & 0.700 & 0.560 & 0.517 & 0.300 & 0.947 & 0.833 & 0.628 & 0.719\\
        LLaMA3-70B& 0.911 & 0.640 & 0.990 & 0.883 & 0.853 & 0.977 & 0.970 & 0.675 & 0.867 & 0.953 & 0.867 & 0.660 & 0.819\\
        \midrule
        \textbf{Maximum}& 0.911 & 0.900 & 0.990 & 0.892 & 0.853 & 0.977 & 0.970 & 0.675 & 0.867 & 0.960 & 0.867 & 0.660 & 0.819\\
        \textbf{Minimum}& 0.232 & 0.530 & 0.540 & 0.308 & 0.500 & 0.000 & 0.000 & 0.092 & 0.000 & 0.427 & 0.342 & 0.356 & 0.396\\
        \bottomrule
    \end{tabular}}
    \caption{The detailed results for \ourdata of all evaluated models. The complete task names are: Dictionary Search, String Completion, Order Check, Order Adjustment, Duplication Check, De-Duplication, Relation Analysis, Count and Navigation, Format Check, Format Cloning, Format Conversion, and List Mapping.}
    \label{tab:detailed-results}
\end{table*}

\begin{table*}[!t]
    \centering
    \small
    \setlength\tabcolsep{1mm}{
    \begin{tabular}{lcccccccccc}
        \toprule
        \multirow{2}{*}{\textbf{Prompt Type}} & \multicolumn{2}{c}{\textbf{Input-Output}} & \multicolumn{2}{c}{\textbf{Question-Answer}} & \multicolumn{2}{c}{\textbf{123-456}} & \multicolumn{2}{c}{\textbf{Q-A}} & \multicolumn{2}{c}{\textbf{answer-question}} \\
        & normal & transfer & normal & transfer & normal & transfer & normal & transfer & normal & transfer \\
        \midrule
        LLaMA2-7B (6-shot)	&0.02	&0.30	&0.23	&0.20	&0.28	&0.23	&0.30	&0.27	&0.18	&0.13 \\
        LLaMA2-13B (6-shot)	&0.00	&0.23	&0.42	&0.23	&0.38	&0.23	&0.40	&0.27	&0.28	&0.20 \\
        \midrule
        LLaMA2-7B (24-shot)	&1.00	&0.98	&1.00	&0.98	&1.00	&0.98	&1.00	&1.00	&1.00	&0.97 \\
        LLaMA2-13B (24-shot)	&1.00	&1.00	&1.00	&1.00	&1.00	&1.00	&1.00	&1.00	&0.97	&0.97 \\
        \bottomrule
    \end{tabular}}
    \caption{Different prompts and n-shot settings in Format Checking Task. "normal" and "transfer" are two different subsets in this task.}
    \label{tab:adjust-prompt}
\end{table*}

\begin{table*}[!t]
    \centering
    \small
    \setlength\tabcolsep{1.2mm}{
    \begin{tabular}{lccccccccc}
        \toprule
        \multirow{2}{*}{\textbf{N-shot}} & \multicolumn{4}{c}{\textbf{LLaMA2-7B}} & \multicolumn{4}{c}{\textbf{LLaMA2-13B}} & \multirow{2}{*}{\textbf{Mean}}\\
	& count-1dim & count-2dim & nav.-1dim & nav.-2dim & count-1dim & count-2dim & nav.-1dim & nav.-2dim	\\
        \midrule
        4-shot	&0.367	&0.067	&0.233	&0.033	&0.500	&0.067	&0.267	&0.033	&0.196 \\
        8-shot	&0.467	&0.067	&0.267	&0.133	&0.567	&0.200	&0.400	&0.067	&0.271 \\
        16-shot	&0.567	&0.000	&0.467	&0.067	&0.833	&0.100	&0.567	&0.033	&0.329 \\
        \bottomrule
    \end{tabular}}
    \caption{Different n-shot settings in Count \& Navigation Task. "count-1dim", "count-2dim", "navigation-1dim", and "navigation-2dim" are four different subsets in this task.}
    \label{tab:adjust-n-shot-count}
\end{table*}


\end{document}